\documentclass[pmlr]{jmlr}

\usepackage{enumitem}

\usepackage{bm} 
\usepackage{longtable}
 \usepackage{booktabs}
 \usepackage{longtable}
\usepackage{hyperref}[colorlinks=true,linkcolor=blue]
\usepackage{url}

\usepackage{wrapfig}
\usepackage{changepage} 
\usepackage{enumitem}
\newcommand{\Xb}{\mathbf{X}}
\newcommand{\Yb}{\mathbf{Y}}
\newcommand{\xb}{\mathbf{x}}
\newcommand{\yb}{\mathbf{y}}
\newcommand{\RR}[0]{\mathbb{R}}
\newcommand{\Gb}{\mathbf{Q}}
\newcommand{\Qb}{\mathbf{G}}

\newcommand{\cL}{\mathcal{L}}
\usepackage{graphicx} 

\usepackage{float}
\usepackage{booktabs}
\usepackage{color}
\usepackage{graphicx}
\usepackage{caption}
\usepackage{float}
\usepackage{array}
\usepackage{makecell}
\usepackage{multirow}
\usepackage{amssymb}
\usepackage{makecell}
\usepackage{booktabs}
\usepackage[load-configurations=version-1]{siunitx} 
\makeatletter
\def\set@curr@file#1{\def\@curr@file{#1}} 
\makeatother
\theorembodyfont{\upshape}
\theoremheaderfont{\scshape}
\theorempostheader{:}
\theoremsep{\newline}

 \usepackage{enumitem}
\setlist[itemize]{noitemsep}

\usepackage[load-configurations=version-1]{siunitx} 

\makeatletter

\newcommand{\equal}[1]{{\hypersetup{linkcolor=black}\thanks{#1}}}

\theorembodyfont{\upshape}
\theoremheaderfont{\scshape}
\theorempostheader{:}
\theoremsep{\newline}

\jmlrvolume{182}
\jmlryear{2022}
\jmlrworkshop{Machine Learning for Healthcare}

\title[ Latent Temporal Flows for Multivariate Analysis of Wearables Data]{Latent Temporal Flows for Multivariate Analysis of Wearables Data}

\author{\Name{Magda Amiridi}\equal{Work done while at Apple.}
       \Email{ma7bx@virginia.edu}\\
       \addr Department of Electrical and Computer Engineering\\
       \addr University of Virginia\\
       \addr Charlottesville, VA, USA
       \AND
       \Name{Gregory Darnell}
       \Email{gdarnell@apple.com}\\ 
       \addr Apple\\
       \addr Cupertino, CA, USA
       \AND
       \Name{Sean Jewell}
       \Email{sean\_j@apple.com}\\
       \addr Apple\\
        \addr Cupertino, CA, USA
   } 

\begin{document}

\maketitle

\begin{abstract}
Increased use of sensor signals from wearable devices as rich sources of physiological data has sparked growing interest in developing health monitoring systems to identify changes in an individual’s health profile. Indeed, machine learning models for sensor signals have enabled a diverse range of healthcare related applications including early detection of abnormalities, fertility tracking, and adverse drug effect prediction. However, these models can fail to account for the dependent high-dimensional nature of the underlying sensor signals. In this paper, we introduce Latent Temporal Flows, a method for multivariate time-series modeling tailored to this setting. We assume that a set of sequences is generated from a multivariate probabilistic model of an unobserved time-varying low-dimensional latent vector. Latent Temporal Flows simultaneously recovers a transformation of the observed sequences into lower-dimensional latent representations via deep autoencoder mappings, and estimates a temporally-conditioned probabilistic model via normalizing flows. Using data from the Apple Heart and Movement Study (AH\&MS), we illustrate promising forecasting performance on these challenging signals. Additionally, by analyzing two and three dimensional representations learned by our model, we show that we can identify participants’ $\text{VO}_2\text{max}$, a main indicator and summary of cardio-respiratory fitness, using only lower-level  signals. Finally, we show that the proposed method consistently outperforms the state-of-the-art in multi-step forecasting benchmarks (achieving at least a $10\%$ performance improvement) on several real-world datasets, while enjoying increased computational efficiency.
\end{abstract}

\begin{figure}
\begin{center}
\includegraphics[scale=0.22]{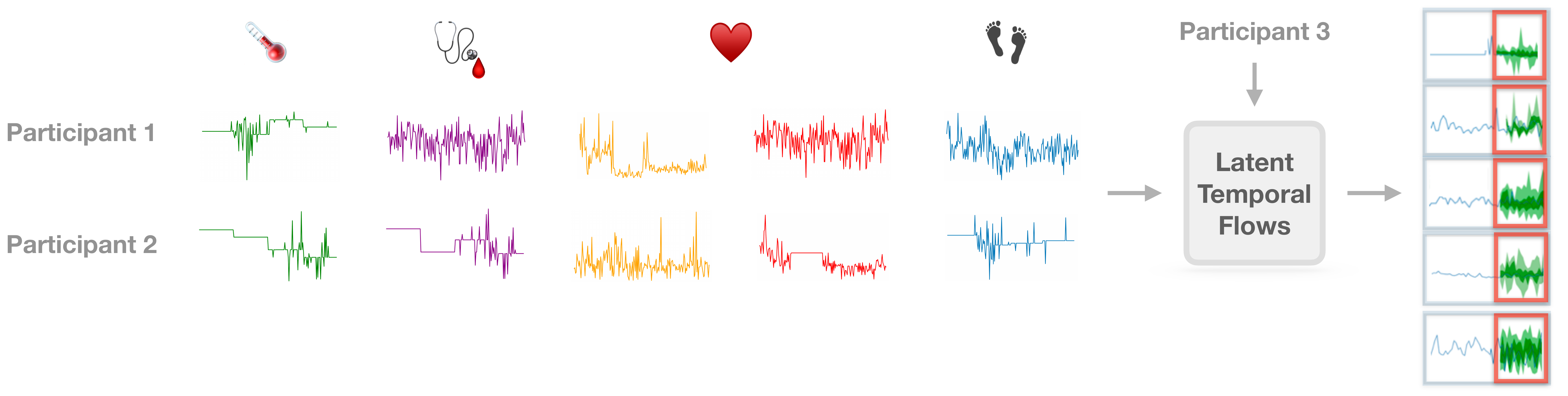}
  \caption{\textbf{Latent Temporal Flows models the joint trajectory of vital signals collected from wearables}: The model is trained to produce $20$-day forecasts of vital signals, given $30$-day multivariate observations of daily heart, blood pressure, and physical activity measurements, collected by wearable devices and manual logging from $1630$ healthy individuals. Each participant's set of sensor measurements is segmented into $50$ day windows, further subdivided into two parts: the context window ($30$ days) and the prediction window ($20$ days). Once the model is trained, given at least $30$ past days of a specified participant in the testing set, we produce multivariate forecasts that capture changes over the vital signals through time.}
   \label{maples_mot}
\end{center}  
\end{figure}
\section{Introduction}
The ubiquity of time-series data in real-world systems, such as healthcare, has created unprecedented opportunities to apply machine learning tools that accurately model complex dynamics to improve individual and population health. For example, multivariate sequential data collected from wrist-worn devices (e.g. heart rate and respiratory rate) can be used to build insights into an individual’s health; complex interactions in sensor signals have been shown to be associated with changes in physiology \citep{goodale2019wearable, natarajan2020assessment, mishra2020pre, radin2020harnessing, radin2021assessment, quer2021wearable, alavi2022real}. 
Joint statistical models of multi-modal signals from wearables can be used as a tool to develop a real-time tracking system for improvement or decline in a patient’s health to facilitate early intervention~\citep{alavi2022real, mishra2020pre}. Another application of interest includes forecasting and analyzing electroencephalogram (EEG) signals, which are critical for identifying emerging dynamic behaviours such as seizures. However, accurate modeling of physiology also requires modeling a large number of (highly-correlated) source signals from different brain regions. 

Expressive time-series models have been used to perform sequential data analysis and forecasting. Specifically, such models allow for long-term predictions with uncertainty estimation \citep{futoma2016predicting}, time-series data augmentation \citep{wen2020time}, out-of-distribution detection \citep{serra2019input}, missing data imputation \citep{luo2018multivariate}, and improved model interpretability analyses \citep{ismail2020benchmarking,rooketemporal}. Furthermore, such information can be integrated into health monitoring systems \citep{dunn2021wearable}.

Given a sequence of data of dimensionality $N$, with $N$ often large, a multivariate time-series forecasting model aims to effectively predict the future outcomes of each variable $n\in \{1, \ldots, N\}$, given past values. Typically, this process requires an accurate estimate of a multivariate conditional distribution of temporal transitions. This task is challenging  in the high-dimensional setting in terms of computational complexity and expressiveness due to the number of parameters required to be estimated. Thus, directly estimating the full predictive distribution of the future values in the high-dimensional observation space is sample-inefficient and may require a large amount of training data.  

The intractability of estimating such models for large systems, due to the growth in the number of parameters with dimension, has limited existing methods to handle at most a few dimensions or requires restrictive assumptions such as tractable distribution classes \citep{chung2015recurrent} or low-rank approximations \citep{salinas2019high}. An alternative way to tackle time-series generative modeling is to view the collection of data as $N$ separate time sequences and fit a separate model for each of the dimensions. However, univariate techniques~\citep{zhang2017stock,oreshkin2019n,montero2020fforma} do not benefit from joint learning of temporal dependencies between features, and are therefore limited in forecasting accuracy. Since many multivariate time-series in practical applications are highly correlated, it is crucial to learn both intra-series and inter-series patterns.  Modeling dependency relationships among the variables  between the individual time-series, including indirect relationships through shared latent causes \citep{wang2019deep}, shows promise in enhancing the performance of forecasting.

In practice, inter-series dependencies may be hidden in high dimensions due to high variance, and thus may be missed when conducting direct data analysis in the higher dimensional space. In such cases, we aim to learn dynamics in a compact latent space to enable faster forecasting, and potentially reveal latent trends. We assume that the observed, possibly high-dimensional, random vector representing the data of interest, is generated from an unobserved low-dimensional latent vector through a time-varying probabilistic process \citep{zhang2019solar, laumer2020deepheartbeat, louis2019riemannian, chung2015recurrent, yoon2019time, nguyen2021temporal}.

By modeling a lower-dimensional latent vector by a (stationary) reversible embedding mapping, Latent Temporal Flows increase parsimony, while reducing computational complexity. These mappings are trained to favor latent representations that uncover latent dynamics over future states.
Under the Latent Temporal Flows framework, forecasts are generated by estimating the latent state evolution, along with recovery functions. Nonlinear mappings are trained via deep autoencoder networks, while latent transition dynamics are learned using a combined structure of Normalizing Flows \citep{rasul2020multi, dinh2016density, papamakarios2017masked} and a multivariate recurrent neural network (RNN) \citep{Graves13, NIPS2014_5346}.

Our experiments on real-world datasets illustrate the state-of-the-art performance of Latent Temporal Flows (at least a $10\%$ performance improvement) and computational tractability on the datasets considered. To demonstrate the ability of the proposed model to match the ground truth trajectory distribution, we apply Latent Temporal Flows to  the Apple Heart and Movement Study (AH\&MS) dataset for a challenging sensor-signal forecasting task. We visualize the 2-dimensional and 3-dimensional learned latent representations from our model to showcase the interpretable low-dimensional embeddings. These emebeddings show that Latent Temporal Flows is able to easily identify participants' $\text{VO}_2\text{max}$, a main indicator and summary of cardiorespiratory fitness, while only being trained on relatively lower-level vital signals like resting heart rate and heart rate variability.

\section{Related work and our contributions}

The literature on time-series forecasting has a long history. In this work, we mainly focus on recent developments in the deep learning context. Simple models for multivariate data, such as general State-Space Models (SSMs) ~\citep{hamilton2020time, liu2016online},  N-BEATS~\citep{oreshkin2019n}, Gaussian Processes ~\citep{rasmussen2003gaussian}, DeepAR~\citep{salinas2020deepar, zhang2017stock} learn one model per time-series (univariate methods). As a consequence, they cannot effectively capture complex structure and interdependencies between multiple time-series.

Although multivariate probabilistic time-series forecasting models estimate the full predictive distribution, the number of parameters in these models grows quickly with the number of variables, which results in a large computational cost and a high risk of overfitting in high dimensional settings. Methods such as Variational Recurrent Neural Networks (VRNNs)  \citep{chung2015recurrent} or Time-GAN~\citep{yoon2019time}, either assume pre-selected tractable distribution classes or another type of structural approximation~\citep{salinas2019high}. In the low-rank Gaussian copula model, for instance~\citep{salinas2019high}, a multitask univariate LSTM~\citep{hochreiter1997long} is used to output transformed time-series and diagonal low-rank factors of a Gaussian covariance matrix. These assumptions can limit the distributional expressiveness of low-rank Gaussian copula models.

Recently, more flexible models such as Temporal Conditioned Normalizing Flows~\citep{rasul2020multi} have been proposed. This method uses a multivariate RNN to learn temporal dynamics  with the state translated to the output joint distribution via  Normalizing Flows \citep{dinh2016density}. However, during forecasting, an invertible flow is applied on the same number of latent dimensions as input dimensions, thus it does not scale to large numbers of time-series (since RNNs have quadratic complexity in $N$). Recent related work also includes Temporal Latent Autoencoder (TLAE), where the temporal model is applied across a low dimensional space~\citep{nguyen2021temporal}. This method combines an RNN-based model with autoencoders to learn a temporal deep learning latent space forecast model. TLAE focuses on the encoder/decoder modeling capabilities and assumes a simplistic probabilistic structure on the latent vector (multivariate Gaussian with diagonal covariance matrix), which can be restrictive.

Our key contributions can be summarized as follows:
\begin{itemize}
       \item     Latent Temporal Flows demonstrates {\it improved forecasting performance while scaling to very high dimensions.}  Our empirical analysis indicates that by projecting input data into an intermediate (much lower dimensional) latent space that preserves most of the essential information (cross-series dependencies), one can obtain improved forecasting performance as a result of the denoising benefits of compact representations.
       \item {\it Efficient training/testing:} In contrast to models operating directly on the high dimensional input space \citep{salinas2019high, rasul2020multi},  fewer parameters in the generative model, as a consequence of reduced dimensionality of the input vectors, leads to a faster training process.  Forecasting in Latent Temporal Flows is performed across a low dimensional space, {\it enabling faster sequence generation.}  In applications such as video prediction, where frame speed generation is critical, training a fast time-series generative model is crucial.  
         \item  {\it Expressivity:} The expressivity of our approach for multivariate forecasting is partly motivated by the fact that we explicitly learn the latent space structure through an expressive time-conditioned distribution model of temporal transitions without strong assumptions of traditional multivariate models (low-rankness, independent components etc.).
    \item  We introduce an {\it end-to-end training process based on a novel cost function.} We are able to harness stochastic gradient descent by combining the objectives for sequence reconstruction and latent probabilistic prediction.
    \item {\it Applicability in forecasting bio-signals}.  We show that Latent Temporal Flows can accurately model the joint trajectory of vital signals collected from wearables (Figure \ref{maples_mot}). Other than improved forecasting performance on the AH\&MS dataset, we visualized the 2-dimensional and 3-dimensional learned latent representations from our model to showcase interpretability in the generated latent representations. These embeddings show that our model is able to easily separate levels of participants’ $\text{VO}_2\text{max}$ (i.e., cardiorespiratory fitness), while only being trained on relatively lower-level vital signals like resting heart rate and heart rate variability.
\end{itemize}

\section{Problem Statement and Proposed Approach}

Consider a collection of high dimensional multivariate time-series $\yb_{n,t} \in \RR$, where $n \in \{1, 2, \ldots, N\}$ indexes the individual univariate component time-series, and $t$ indexes time. Consequently, the multivariate observation vector at time $t$ is given by $\yb_t \in \RR^N$. Given a sequence of $T_\text{total}$ vector realizations of $\yb_t$, a multivariate time-series can be represented as a matrix $\Yb$, $ \Yb \in \RR^{N \times T_\text{total}}$. We focus on the task of multivariate time-series multi-step forecasting. More formally, let us assume that we are given an observed history $\big(\yb_{1}, \ldots, \yb_{T}\big)$,  sampled from the complete time-series history of the training data,  where each instance consists of $N$ temporal features (that occur over time, e.g. vital sensor signals). Our goal is to learn future values of the series over a length-$\tau, \tau>0,$ forecast horizon and predict a set of plausible future trajectories $\big(\hat{\yb}_{T+1}, \ldots, \hat{\yb}_{T+\tau}\big)$  by learning the conditional distribution of temporal transitions,
\begin{equation*}
    p(\yb_{T+1:T+\tau} | \yb_{1:T}).
\end{equation*}
We refer to time-series $\yb_{n,1:T+\tau}$ as the target time-series, which for training is split according to a time range $\big(1,2,\ldots,T\big)$ referred to as a context window, and to time $\big(T+1,T+2,\ldots,T+\tau\big)$ as a prediction horizon.

The basic idea of the proposed model can be described as follows. The temporal dynamics of complex systems are often driven by fewer and lower-dimensional factors of variation \citep{laumer2020deepheartbeat, louis2019riemannian, chung2015recurrent,yoon2019time, nguyen2021temporal, amiridi2021low}. We assume that the observed, possibly high-dimensional, random vector $\yb_t \in \RR^{N}$ representing the data of interest is generated from an unobserved low-dimensional latent vector $\xb_{t} \in \RR^{D}$ with $D \ll N$ through a time varying probabilistic process $p(\xb_{t}|\xb_{1:{t-1}})$. To reduce the computational burden while at the same time improving distribution modeling, we propose incorporating representation learning in the generative learning problem to explicitly learn the temporal distribution of compact representations of input sequences. Specifically, Latent Temporal Flows consists of three key components: an embedding function $g: \RR^N \rightarrow \RR^D$ and a recovery function $q: \RR^D \rightarrow \RR^N$, which are learned via an autoencoder neural network, as well as a latent conditional distribution $p(\xb_{t}|\xb_{1:{t-1}})$ of temporal transitions, modeled by a time conditioned Normalizing Flow. During training, the model simultaneously learns to produce sequence representations that will push a latent temporal generative model to predict plausible latent future sequences, recover them back into the observed space, and iterate across time. 
 
 Given a model for $p(\yb_{T+1:T+\tau} | \yb_{1:T} )$, one can estimate the conditional expectation, which can be expressed as a function of past observations: $\mathbb{E}[\yb_{T+1:T+\tau} | \yb_{1:T} ]= f(\yb_{1}, \ldots, \yb_{T}).$ An indirect estimate of this function can be accomplished using the following strategy: Future trajectories $\big(\hat{\yb}_{T+1}, \ldots, \hat{\yb}_{T+\tau}\big)$ can be generated by first embedding the past history via 
  $g: \big({\xb}_{1}=g(\yb_1), \ldots, {\xb}_{T}=g(\yb_{T})\big),
$ sampling  from the latent temporal distribution 
  $ \big(\hat{\xb}_{T+1}, \ldots, \hat{\xb}_{T+\tau}\big) \thicksim p(\cdot | \xb_{1:T}),$ followed by applying a non-linear recovery function $q$: $\big(\hat{\yb}_{T+1}=q(\hat{\xb}_{T+1}), \ldots, \hat{\yb}_{T+\tau}\}=q(\hat{\xb}_{T+\tau})\big).
$ Next, we describe each building block of our approach and the combined training strategy.

\subsection{Learning Sequence Representations}

 The goal of the proposed method is to obtain suitable representations  $\xb_t \in \RR^{D}$ that reveal a reduced search space for future sequence forecasting, where underlying patterns and meaningful information among features is preserved.  Latent-space forecasting has been considered both  in~\citet{yu2016temporal} as a result of matrix factorization as well as in \citet{nguyen2021temporal}. In~\citet{yu2016temporal}, a multivariate time-series  $\Yb$ is decomposed into components $\Gb \in \RR^{N \times D} \text{ and } \Xb \in \RR^{D \times T_\text{total}}$, with $\Xb$ temporally constrained. Let us denote $\Qb$ as the pseudo-inverse of $\Gb$. If $\Yb$ can be decomposed by $\Qb$ and $\Xb$, forecasting for the high-dimensional series $\Yb$ can be performed on a smaller dimensional series. In TLAE \citep{nguyen2021temporal}, linear mappings are generalized to nonlinear transformations via time-varying autoencoders. In our framework, each input vector $\yb_t \in \RR^{N}$ is mapped to a condensed representation $\xb_t \in \RR^{D}$ (usually $D \ll N$) using a \emph{stationary} reversible embedding mapping ${g}:\mathbb{R}^N \rightarrow \mathbb{R}^D$. Such representations are trained to reveal a low-dimensional structure, which allows an expressive family of temporal constrained  distributions to fully uncover this, i.e., mappings that yield a low negative log-likelihood cost over $p(\xb_{t}|\xb_{1:{t-1}})$.

The first component of our model is estimating the dimensionality reducing mapping. Exploiting the deep neural network’s ability to to extract higher order features \citep{yang2017towards} and approximate any nonlinear function,  we replace $\Qb$ by an encoder and $\Gb$ by a decoder neural network. An encoder network $g_{\bm \phi}: \RR^N \rightarrow \RR^D$, embeds $\yb_t$ at time $t$ into a low-dimensional latent space, vector $\xb_t$, using a nonlinear map.  Operating within the latent space, we then seek, a dynamical system that prescribes a rule to move forward in time, and a decoder network 

\begin{equation}
q_{\phi'}: \RR^{D}: \rightarrow \RR^N, \hat{\yb}_t = q_{\bm \phi'}(\xb_t)
\end{equation}
 to reconstruct latent variables in the spatial domain. Although the embedding mapping is assumed to be stationary, latent representations' progressions over time are captured via an autoregressive deep learning model, where the data distribution is represented by a conditioned Normalizing Flow. During training, the autoencoder learns by fine-tuning the parameters of a feed-forward Deep Neural Network (DNN) in such a way that the reconstruction error is minimized when back projected with another feed-forward DNN. These networks need to be specified a-priori, in terms of the number of layers and neurons. We note that the specific requirements for $g$ and $q$ are problem dependent, and we detail the particular design we use in the Appendix. Given a batch of time-series $\mathcal{B}$, with $|\mathcal{B}|$ denoting the cardinality of the batch set, the first term of our overall cost function consists of the reconstruction loss: 

\begin{equation}
     {\mathcal{L}_{\text{REC}}}:=\frac{1}{|\mathcal{B}|(T+\tau)} \sum_{
\mathbf{y}_{1:T+\tau} \in \mathcal{B}}\sum_{t=1}^{T+\tau} {{||{\mathbf y}_t-{q}({g}({\mathbf y}_t;{{\phi}});{{\phi'}})||}_2^2}.
\end{equation}

\subsection{Compressed Sequence Modeling}
The latent random vector lies  at the heart of our overall probabilistic model: It is assumed to ``encode'' the observed data in a compact manner through  $\xb_{t} = g_{\bm \phi}(\xb_{t})$, allowing an accurate model of the probabilistic model $p_\theta(\xb_{t}|\xb_{1:{t-1}})$, from which new data can be generated. Our goal is to recover the latent space structure through a flexible time-conditioned distribution model in which the most important features are kept. We also learn the mapping $\hat{\xb_t} = q_{\bm \phi'}(\xb_t)$ that translates the latent effects to the original data space.

Using the chain rule, the joint distribution of predicted values conditioned on observed values, $p(\xb_{T+1}, \ldots, \xb_{T+\tau} | \xb_{1:T})$, can be written as a product of conditional distributions. Autoregressive models use a neural network to approximate the conditional distribution $p(\xb_{t} | \xb_{1:t-1})$ by a parametric distribution $p_\theta(\xb_{t} | \xb_{1:t-1})$ specified by learnable parameters $\theta$. The prediction at time $t$ is input to the model to predict the value at time $t+1$:

\begin{equation}
p(\xb_{T+1}, \ldots, \xb_{T+\tau} | \xb_{1:T} ) = \prod_{t=T+1}^{T+\tau} p(\xb_{t} | \xb_{1:t-1} ).
\end{equation}

  We wish to replace this decomposition  by a tractable, approximate statistic $\mathbf{h}_t$ of the past. To represent the history of observations in a compressed state vector $\mathbf{h}_{t}$ we use RNNs~\citep{Graves13, NIPS2014_5346}, with the most well-known variants, the LSTM~\citep{hochreiter1997long} and GRU~\citep{69e088c8129341ac89810907fe6b1bfe}.

We assume that a time-related (and recursively updated) vector $ \mathbf{h}_{t}\in \RR^{H}$, can summarize the history of the time-series up to time time point $t$, $\mathbf{h}_t = \mathrm{RNN}_\theta(\xb_{t-1}, \mathbf{h}_{t-1})$,
where $\mathrm{RNN}_\theta$ is a multi-layer LSTM or GRU parameterized by shared weights $\theta$ and $\mathbf{h}_0 = \mathbf{0}$. The state is compressed as it uses less space than the history of observations. Under this model, we can factorize the joint distribution of the observations as $(\xb_{T+1}, \ldots, \xb_{T+\tau} | \xb_{1:T} ) = \prod_{t=T+1}^{T+\tau}   p_\theta(\xb_{t} | \mathbf{h}_{t})$, where now $\theta$ comprises both the weights of the RNN as well as the probabilistic model. This model is auto-regressive as it consumes the observations at the time step $t-1$ as input to learn the distribution of the next time step $t$. Then, time conditioning on the latent generative model $p_\theta(\xb_t|\mathbf{h}_{t})$ can be realized by employing a multivariate RNN to model the series progressions, with the state translated to the output joint distribution via a flow (we focus on Real-NVP, but MAF is also explored in our experiments; background on Normalizing Flows can be found in the following paragraphs). This combination retains the power of autoregressive models---such as good performance in extrapolation into the future---with the flexibility of flows as an expressive distribution model. 

\paragraph{Background on Normalizing Flows:}
\emph{Normalizing Flows}~\citep{rezende2015variational,dinh2014nice, dinh2016density, papamakarios2017masked,kingma2018glow, chen2019residual, papamakarios2019normalizing} define a smooth, invertible transformation  $f \colon \mathcal{X} \mapsto \mathcal{Z}$, of a simple base distribution $p(\mathbf{z})$ (e.g. an isotropic Gaussian) on the space $\mathcal{Z}=\mathbb{R}^{D}$ into a more complex  distribution $p(\mathbf{x})$ on the space $\mathcal{X}=\mathbb{R}^{D}$ by a sequence of invertible and differentiable mappings. Its reverse operation $\mathbf{x} = f^{-1}(\mathbf{z})$ synthesizes realistic samples from the prior, is easy to evaluate, and computing the Jacobian determinant takes $\mathcal{O}(D)$ time. Via the change of variables formula, $p(\mathbf{x})$ can be expressed as 
\begin{equation*}
p(\mathbf{x}) = p(\mathbf{z}) \left| \det \left( \frac{\partial f(\mathbf{x})}{\partial \mathbf{x}} \right)\right|.
\end{equation*}

Real-valued non-volume preserving (RealNVP) models introduce a \emph{coupling layer}, which is the building block/bijection that leaves part of its inputs unchanged $\mathbf{c}^{1:d} = \mathbf{x}^{1:d}$ and transforms the other part via functions of the untransformed variables 
    $\mathbf{c}^{d+1:D} =  \mathbf{x}^{d+1:D} \odot \exp(s( \mathbf{x}^{1:d})) + t( \mathbf{x}^{1:d})$.
Here, $\odot$ is an element wise product, $s()$ is a scaling and $t()$ a translation function from $\mathbb{R}^{d} \mapsto \mathbb{R}^{D-d}$, given by  neural networks. To model a nonlinear density map $f(\mathbf{x})$, a number of coupling layers which map $\mathcal{X} \mapsto \mathcal{C}_1 \mapsto \ldots \mapsto \mathcal{C}_{K-1} \mapsto \mathcal{Z}$ are composed together all the while alternating the dimensions which are unchanged and transformed. Via the change of variables formula the probability density function given a data point can be written as
$\log p(\mathbf{x})  = \log p(\mathbf{z}) + \log | \det(\partial \mathbf{z}/ \partial\mathbf{x})| \nonumber= \log p_{\mathcal{Z}}(\mathbf{z}) +  \sum_{i=1}^{K} \log | \det(\partial \mathbf{c}_{i}/ \partial\mathbf{c}_{i-1})|.$

\sloppy Following \cite{rasul2020multi}, we concatenate $\mathbf{h}$ to the inputs of the scaling and translation function approximators of the coupling layers, i.e. $s(\mathtt{concat}(\mathbf{x}^{1:d}, \mathbf{h}))$ and $t(\mathtt{concat}(\mathbf{x}^{1:d}, \mathbf{h}))$.  The model, which is parameterized by both the flow (the weights of the scaling and translation neural networks) and the RNN --  $\theta$ is trained by minimizing:
\begin{equation}
{\mathcal{L}_\text{NegLL}} :=\frac{1}{|\mathcal{B}|(T+\tau)} \sum_{
\mathbf{y}_{1:T+\tau} \in \mathcal{B}}\sum_{t=T+1}^{T+\tau} \log  p(\xb_{t} | \mathbf{h}_{t};\theta).
\end{equation}
\subsection{Latent Temporal Flows: Model Fitting and Inference}
\begin{figure}
 \centering
   \includegraphics[width=5in]{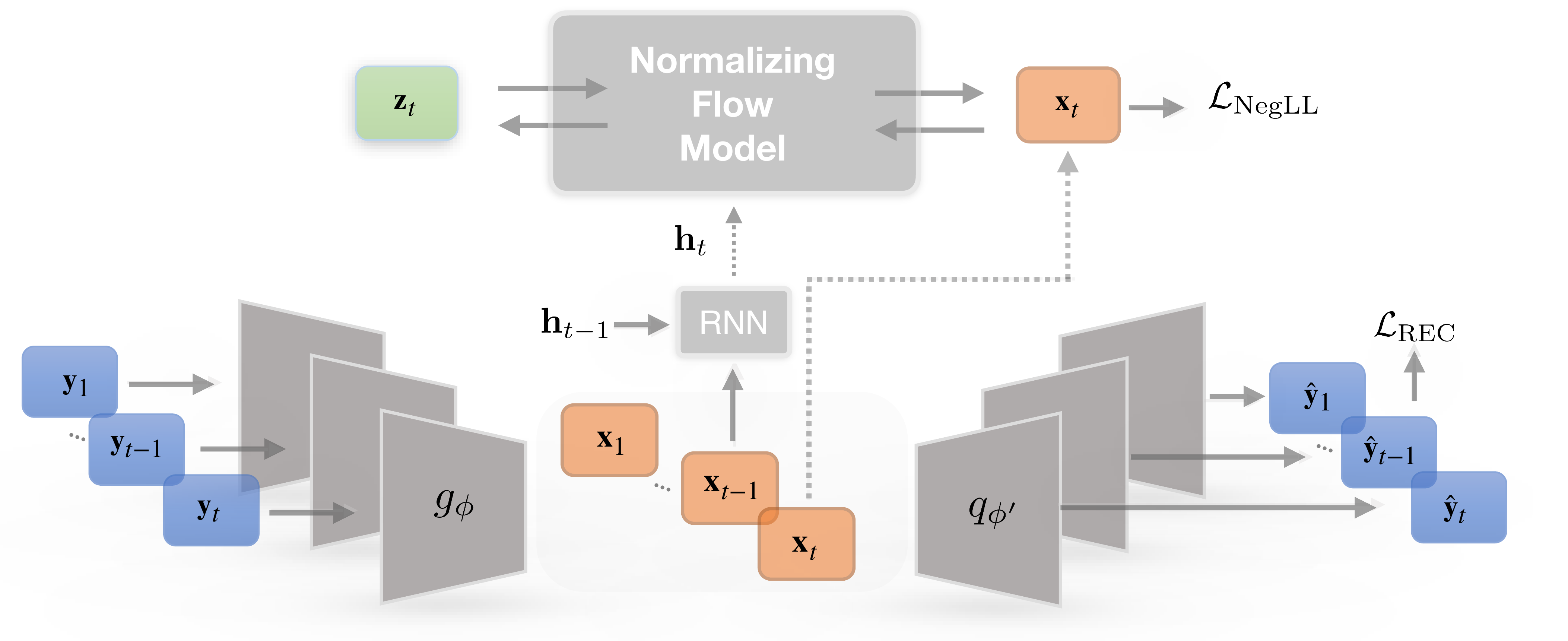}  
            \caption{\textbf{Information flow during the training process}: 
  {  Our model consists of three key components: an embedding mapping ${g}$, a latent conditional distribution of temporal transitions, and a recovery function ${q}$. Mappings ${g}$ and ${q}$ are trained through a deep autoencoder network, while latent transition dynamics are learned using a combined structure of Normalizing Flows and a recurrent neural network. In this way,  the proposed model increases parsimony, while reducing computational complexity.} }
   \label{fig:training} 
   \end{figure}

 A schematic overview of the training procedure is depicted in Figure \ref{fig:training}. Each time-series $\Yb$ is split into a training ${\Yb}_\text{TR}$ and test set ${\Yb}_\text{TS}$ by using all data prior to a fixed date for training and using rolling windows for the test set. A batch input $\Yb_B \in \RR^{D\times (T+\tau)}$ defines a sub-matrix of ${\Yb}_\text{TR}$ with column indices defined by the set $\mathcal{B}$ -- it contains two components: the first part ${\{{\yb}_t\}}_{t=1}^T$ is associated with the past input, while the second component ${\{{\yb}_t\}}_{t=T+1}^{T+\tau}$ is associated with future observations. A batch of time-series $\Yb_B$ is embedded into  latent variables via $g: {\xb}_t = {g}({\mathbf y}_t)$,  yielding a matrix of latent codes $\Xb_B \in \RR^{D\times (T+\tau)}$.  To discover informative representations in a lower-dimensional space, each vector $\xb_t$ is passed through the decoder layers: $\hat{\yb}_t = {q}({g}({\mathbf y}_t)$, yielding a reconstructed matrix $\hat{\Yb}_B$. 
\begin{itemize}
    \item By minimizing the reconstruction error $\|\hat{\yb}_t - \yb_t\|^2_{2}$,  the model is expected to capture feature dependencies across time-series and encode this global information into a few latent variables in $\xb_t$.
    \item  Simultaneously, by maximizing the log-likelihood of (latent) future observations given compressed past input, the model targets capturing latent series progressions via time-conditioned Normalizing Flows. 
\end{itemize}
The proposed model is meaningful as it encourages the latent variables to capture different complex patterns of the data, which makes the representation more powerful and universal. 

The key insight is that the encoding/decoding components and the time conditioned latent temporal generative model are jointly trained by minimizing a combined loss function over a given batch $\mathcal{B}$ of time-series consisting of a reconstruction loss-related term and a sequence negative log-likelihood term using stochastic gradient descent-based optimization. The overall loss term is 
\begin{equation} 
\mathcal{L}(\phi, \phi', \theta) :=\frac{1}{|\mathcal{B}|(T+\tau)} \sum_{
\mathbf{y}_{1:T+\tau} \in \mathcal{B}}\sum_{t=1}^{T}\big{(} \underbrace{{||{\mathbf y}_t-{q}({g}({\mathbf y}_t;{{\phi}});{{\phi'}})||}_2^2}_{\mathcal{L}_{\text{REC}}}-\lambda \underbrace{\log p({q}({g}({\mathbf y}_t;{{\phi}});{{\phi'}}) | \mathbf{h}_t;\theta)}_{\mathcal{L}_\text{NegLL}}\big{)}.
\end{equation}

By minimizing a combined error of two tractable losses $\mathcal{L}_{\text{REC}}$ and $\mathcal{L}_\text{NegLL}$, the model is given the capability to predict the future from latent representations of the observed history that preserve only the essential information, which is transferred to the decoder by minimizing the reconstruction loss. At the same time, it serves to reduce the dimensions of the temporal generative learning space. The overall optimization problem is
\begin{equation}
\label{eq:tlae}
\textstyle
\min_{ \phi, \phi', \theta } \cL( \phi,  \phi', \theta ) = \frac{1}{|\mathcal{B}|(T+\tau)} \sum_{
\mathbf{y}_{1:T+\tau} \in \mathcal{B}}\sum_{t=1}^{T} \mathcal{L}_{\text{REC}}( \phi, \phi')+\lambda \mathcal{L}_\text{NegLL}(\theta),
\end{equation}
where $\lambda\geq0$ is a hyperparameter that balances the two losses.
 
  \begin{figure}
 \centering
   \includegraphics[width=6in]{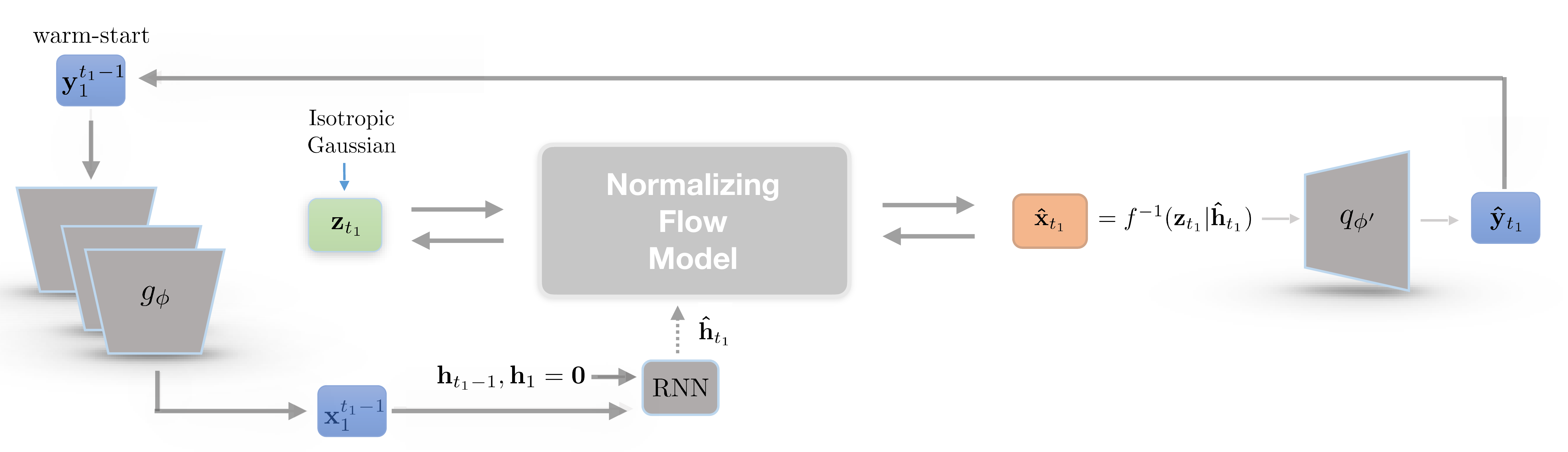} 
    \caption{ \textbf{Forecasting under Latent Temporal Flows }: Given a starting state, the flow generates estimates for next time step by sampling from $\big(\hat{\xb}_{t_1})\thicksim p(\xb_{t_1}|\mathbf{h}_{t_1})$. This process is repeated till our inference horizon. A latent trajectory is mapped back into the input space through the decoder.} 
   \label{Forcast} 
   \end{figure}
Given a trained model, our goal is to produce a set of future trajectories given the past (see Figure~\ref{Forcast}). Note that, instead of producing samples directly in a possibly very high dimensional feature space, the generator first produces Monte Carlo samples from the predictive distribution $p(\xb_{T+1}, \dots \xb_{T+\tau}|\xb_1, \ldots, \xb_T)$ in the lower dimensional embedding space, by sequentially sampling from $p(\xb_t|\mathbf{h}_t)$, updating $\mathbf{h}_t$, and passing the latent sequences through $q_{\bm \phi'}$ to map $\xb_t$ back to the original domain.

\section{Experimental results}

\subsection{Results on AH\&MS dataset}

Biometrics collected from wearable devices, including heart rate measurements and activity levels (e.g., step counts, standing hours) throughout the day, are rich sources of information that can yield crucial insights into the health trajectory of an individual.
Vital signals from wearables show feasibility for accurate prediction of clinical laboratory measurements~\citep{dunn2021wearable}.
These measurements also show promise to detect acute infections---cosinor models fit to diurnal heart rate variability (HRV) patterns can detect pre-symptomatic COVID-19 infection~\citep{hirten2021use}, and anomaly detection algorithms using resting heart rate (RHR) and step counts can identify pre-symptomatic COVID-19~\citep{alavi2021real}.
Biometrics also show sensitivity to detect common colds (H1N1 and rhinovirus)~\citep{grzesiak2021assessment}. Beyond acute infections, wearable measurements can provide insight into cardiovascular health, as higher RHR is significantly associated with coronary artery disease, stroke, and sudden death~\citep{zhang2016association}. All-cause and cardiovascular mortality is also indicated by lower HRV~\citep{singh2018heart}.
HRV is also found to indicate dysregulation of the autonomic nervous system, and thus is implicated in the development of hypertension~\citep{schroeder2003hypertension}.
Joint modeling of vital signs from wearable devices continues to yield powerful signals for early detection of disease and health decline, as well as indicate general well-being and fitness. The relatively low cost of wearables and longitudinal nature of measurements compared to in-clinic evaluations motivates continued research in this area.


\begin{figure}
\begin{center}
  \begin{tabular}{@{}cccc@{}}
            \includegraphics[width=.25\textwidth]{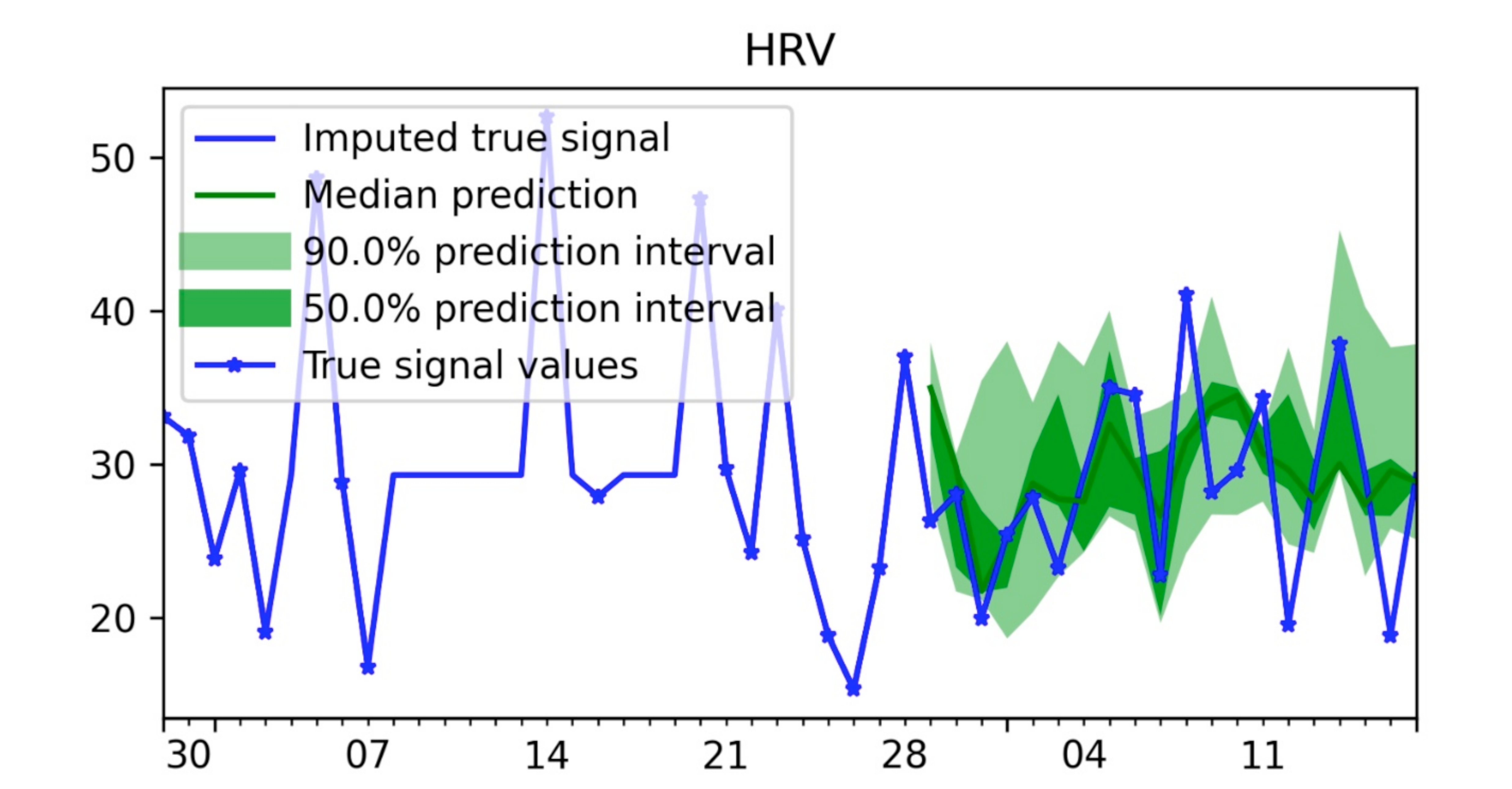}  & 
    \includegraphics[width=.25\textwidth]{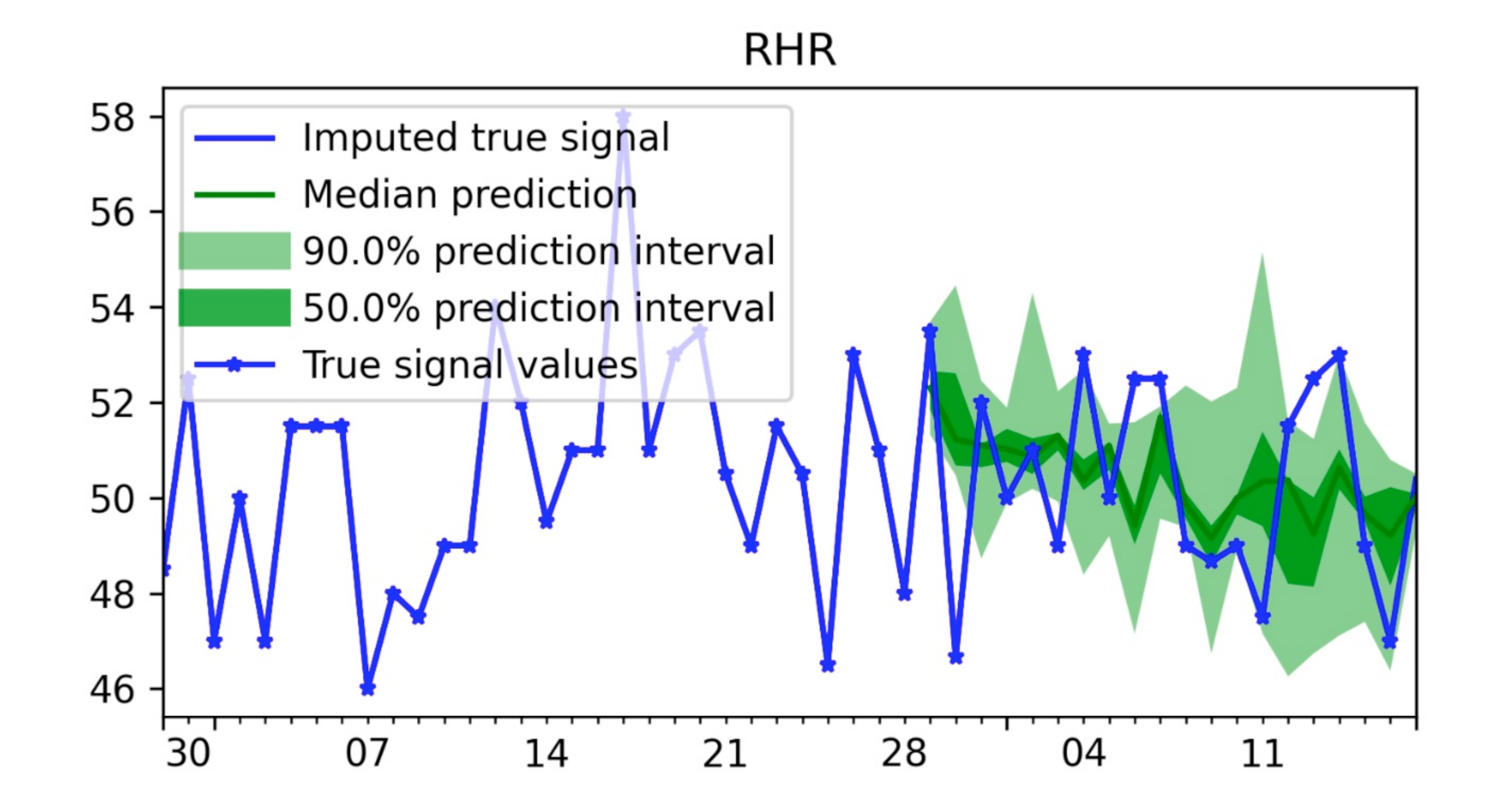} &
    \includegraphics[width=.25\textwidth]{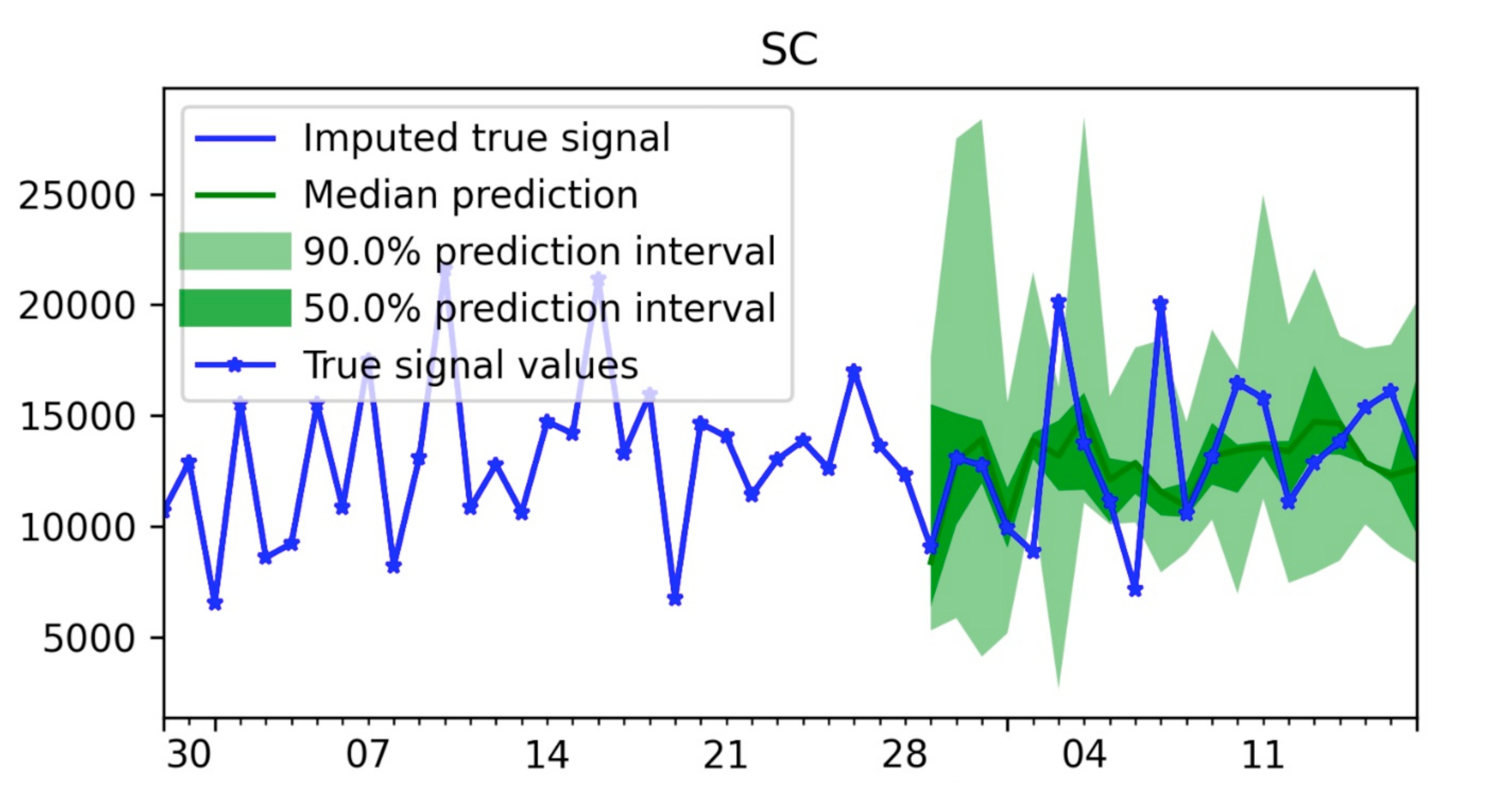} \\
            \includegraphics[width=.25\textwidth]{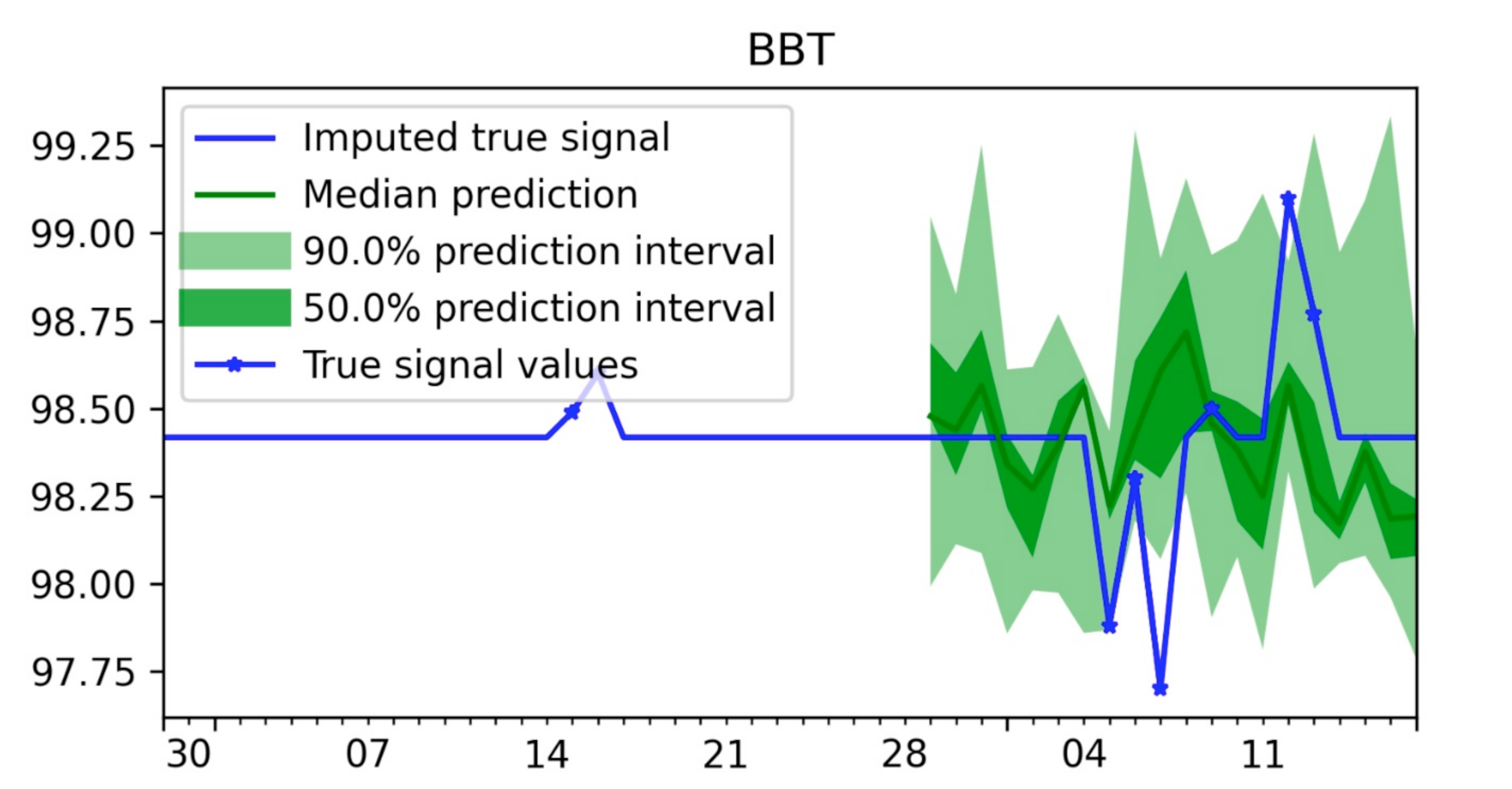} &
    \includegraphics[width=.25\textwidth]{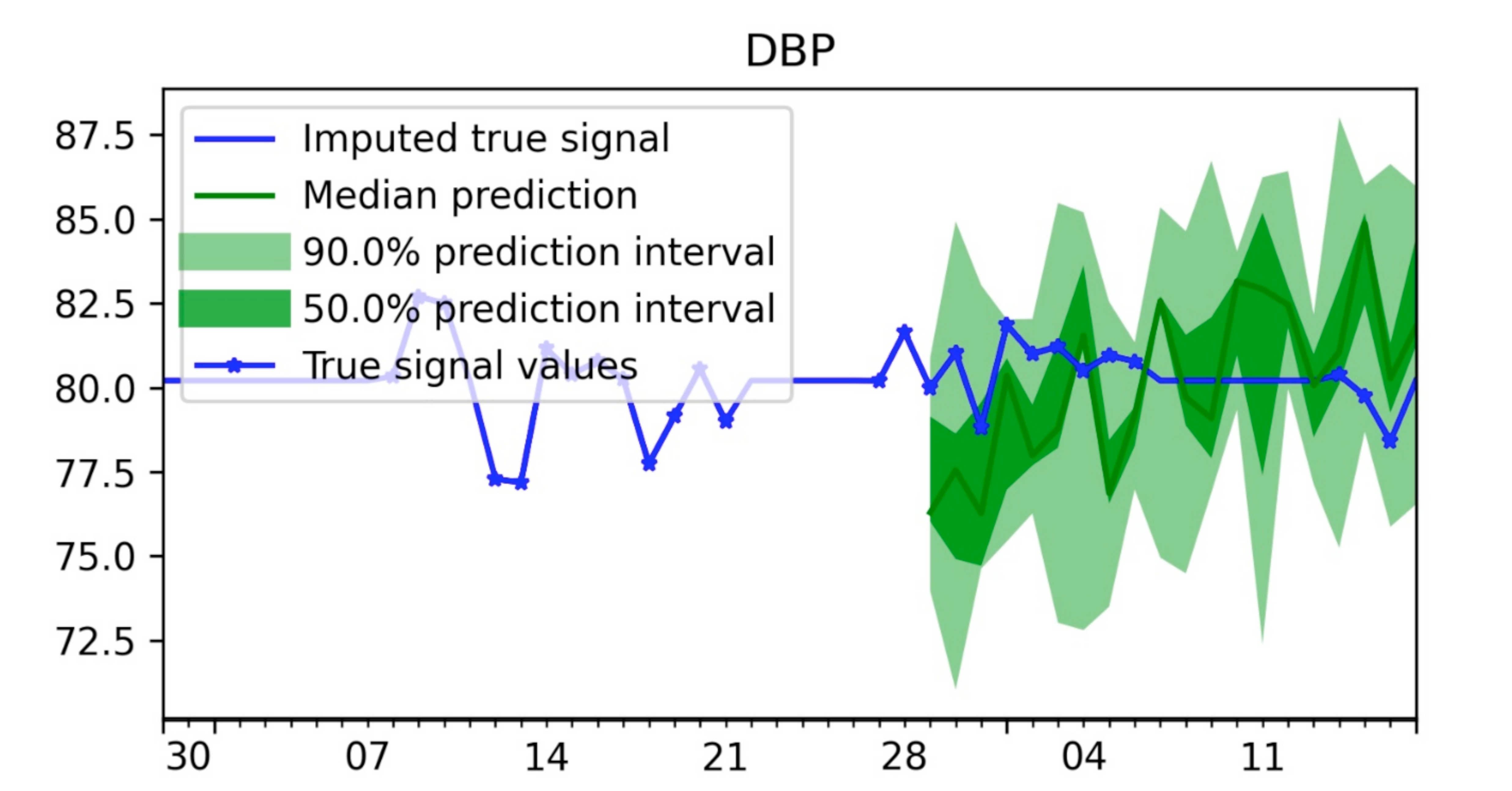} &
    \includegraphics[width=.25\textwidth]{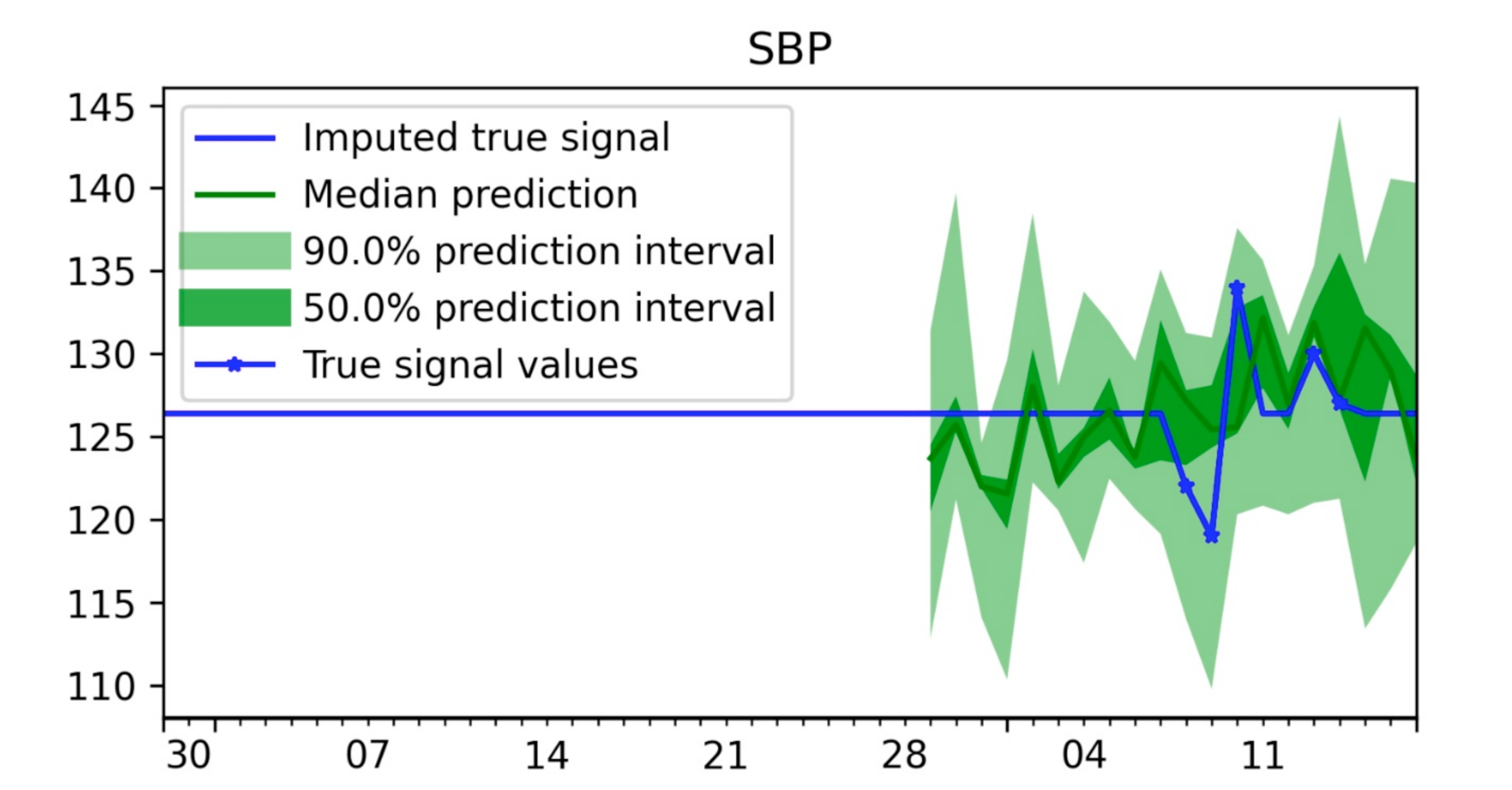} &
      \end{tabular}
  \caption{Qualitative predictions for the $20$ day forecast window produced by Latent Temporal Flows for AH\&MS signal observations (HRV, RHR, SC, BBT, DBP, SBP) of an individual selected at random. We estimate $10$ trajectories for a latent dimension of $D=5$ and plot the estimates. Note that the forecast is displayed in terms of a probability distribution: the shaded areas represent the 50$\%$ and $90\%$ prediction intervals, respectively, centered around the median (dark green line). The ground truth is overlaid in blue with stars denoting true observations and linear interpolation between the points.}
   \label{maples_main}
\end{center}  
\end{figure}

To this target, we apply Latent Temporal Flows {(abbreviated as LatTe)} for multivariate signal forecasting: given the past $30$-day measurements of a group of vital sensor signals as input, we wish to predict the future $20$-day values of the multivariate input time-series. We evaluate the empirical performance of our method on the Apple Heart and Movement Study (AH\&MS) dataset. The AH\&MS study was sponsored by Apple and conducted in collaboration with the American Heart Association and Brigham and Women’s Hospital. The study was approved by Advarra IRB and data were collected in accordance with the IRB approved consent form.

\textbf{AH\&MS dataset description}: The dataset contains signals collected in real-world environments, which constitute a measure of cardiovascular and autonomic nervous system activity as well as movement and activity metrics. For the participants enrolled in this study, this dataset consists of passively collected sensor signal data from wearable devices in addition to self-reported measurements. This rich collection of signals comprises a partial view of individual biometric information as it evolves over time (see Figure \ref{maples_mot}).

Our data consists of passively collected observations: Resting Heart Rate (\textbf{RHR}), Heart Rate Variability (\textbf{HRV}), Step Count (\textbf{SC}), and user logged information such as Diastolic Blood Pressure (\textbf{DBP}), Systolic Blood Pressure (\textbf{SBP}), and Basal Body Temperature (\textbf{BBT}). HRV is calculated in the time-domain as the standard deviation between heartbeat measurements (defined by measured R-R intervals). Each time-series measurement is aggregated at the daily level and averaged if there are multiple measurements per day (for RHR, HRV, DBP, SBP, BBT) or summed (for SC).
\begin{table*}
\begin{center}
\scalebox{0.55}{
\begin{tabular}{llcllll|l|l}
\toprule
 & \makecell{ {NMSE-BBT}} & \makecell{{NMSE-BPD}} & \makecell{{NMSE-BPS}}  & \makecell{{NMSE-HRV}} & \makecell{{NMSE-RHR}} & \makecell{{NMSE-SC}}& \makecell{{CRPS-Sum}} \\
\midrule
{\makecell{{LatTe}, {D=3}}}
&$0.0124\pm 0.0028$
&$0.0038\pm 0.0021$
&$0.0117\pm 0.0464$
&$0.0828\pm 0.0718$
&$0.0145\pm 0.0115$
&$0.3650\pm 0.4169$
&$0.1068\pm 0.0154$
\\

{\makecell{{LatTe}, {D=4}}}
& $0.0114\pm 0.0027$
&$0.0033\pm 0.0018$
&$0.0115\pm 0.0040$
&$\mathbf{0.0593\pm 0.0650}$
&$0.0146\pm 0.0117$
&$0.3264\pm 0.4315$
&$0.1027\pm 0.0156$
\\
{\makecell{{LatTe}, {D=5}}}
&$\mathbf{0.0047\pm 0.0001}$
&$\mathbf{0.0015\pm 0.0011}$
&$\mathbf{0.0089\pm 0.0006}$
&$0.0686\pm 0.0071$
&$0.0073\pm 0.0053$
&$\mathbf{0.3080\pm 0.3061}$
&$\mathbf{0.0841\pm 0.0332}$
\\
\makecell{{DeepVAR}}
&$0.0277\pm 0.0021$
&$0.0092\pm 0.0033$
&$0.0107\pm 0.0006$
&$0.0817\pm 0.0056$
&$0.0129\pm 0.0148$
&$0.4129\pm 0.4823$
&$0.1281\pm 0.0154$
\\

\makecell{{GP} - {Copula}}
&$0.0223\pm 0.0016$
&$0.0062\pm 0.0043$
&$0.0105\pm 0.0056$
&$0.1015\pm 0.0516$
&$0.0427\pm 0.0536$
&$0.3206\pm 0.4152$
&$0.1208\pm 0.0181$
\\

\makecell{{{{TCNF} {LSTM-MAF}}}}
&$0.0225\pm 0.0033$
&$0.0081\pm 0.0031$
&$0.0116\pm 0.005$
&$0.1165\pm 0.0593$
&$0.0718\pm 0.0664$
&$0.3208\pm 0.3246$
&$0.1052\pm 0.0166$
\\
\makecell{ {{TCNF} {Transformer-MAF}}}  
&$0.0224\pm 0.0018$
&$0.0089\pm 0.0024$
&$0.0109\pm 0.0490$
&$0.1094\pm 0.0539$
&$\mathbf{0.0064\pm 0.0226}$
&$0.3144\pm 0.2726$
&$0.0925\pm 0.0155$\\
\bottomrule
\end{tabular}
}

\caption{We present the mean and standard errors obtained by Deep-VAR, GP Copula, TCNF, and our model variants LatTe with Real-NVP. The evaluation criterion is the Test set Normalized Mean Square Error (NMSE) comparison (lower is better). Best performance for each column is highlighted in bold. \label{crps1}}
\end{center}

\vspace{-6mm}
\end{table*}

\begin{figure}
\centering
\includegraphics[width=1\textwidth]{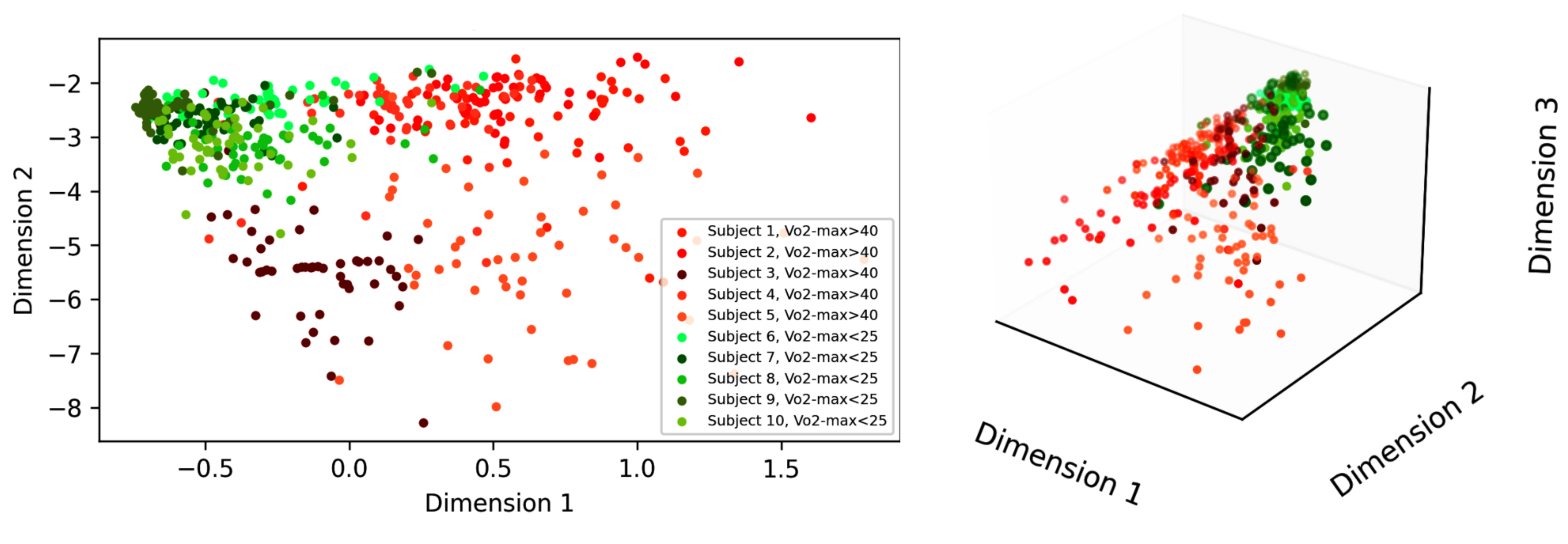} 
  \caption{{\bf The latent manifold of the data embedded in 2D latent space (left) and 3D latent space (right) learned by the autoencoder.} Each point represents a latent signals' representation for a specific time point. Every different color represents a different individual, but those with lower $\text{VO}_2\text{max}$ values are represented in red tones, while those with higher $\text{VO}_2\text{max}$ values in green tones. We observe that learned maps from Latent Temporal Flows clusters subjects according to their estimated fitness level.}
 \label{latent_reps_2d3d}
\end{figure}

\textbf{AH\&MS dataset analysis and results}: Given a training set comprised of multiple context windows of $30$ daily observations of BBT, DBP, SBP, HRV, RHR, and SC measurements of $1630$ individuals, the goal is to jointly learn each signal output for the next $20$ days, by modeling the temporal dependencies of latent representations of the signals while also learning a meaningful lower-dimensional space. We project another signal captured in our dataset, $\text{VO}_2\text{max}$ (omitted from training), onto the learned latent space to show that our model captures meaningful representations (Figure~\ref{latent_reps_2d3d}).

Using the Normalized Mean Square Error (NMSE) on each individual time-series, and CRPS-Sum as an evaluation metric, we qualitatively assess test-time signal predictions produced by our method for different latent-space dimensionality, $D=\{3,4,5\}$. We compare our approach against existing classical multivariate methods. Because of different scales in each signal, we first normalize each signal by the sum of its absolute values before computing this metric. The results are reported in Table~\ref{crps1}, where, on average, Latent Temporal Flows with $D=5$ achieves the best signal predictions. Upon closer examination, we notice that step counts are more difficult to predict compared to measures such as heart rate signals, since step counts are largely influenced by behavior and lifestyle compared with signals like heart rate variability whose variation is significantly affected by less easily modifiable physiological and pathological factors~\citep{fatisson2016influence}.

\begin{figure}
\begin{center}
  \begin{tabular}{@{}cccc@{}}
  \Xhline{2\arrayrulewidth}
            \includegraphics[width=.25\textwidth]{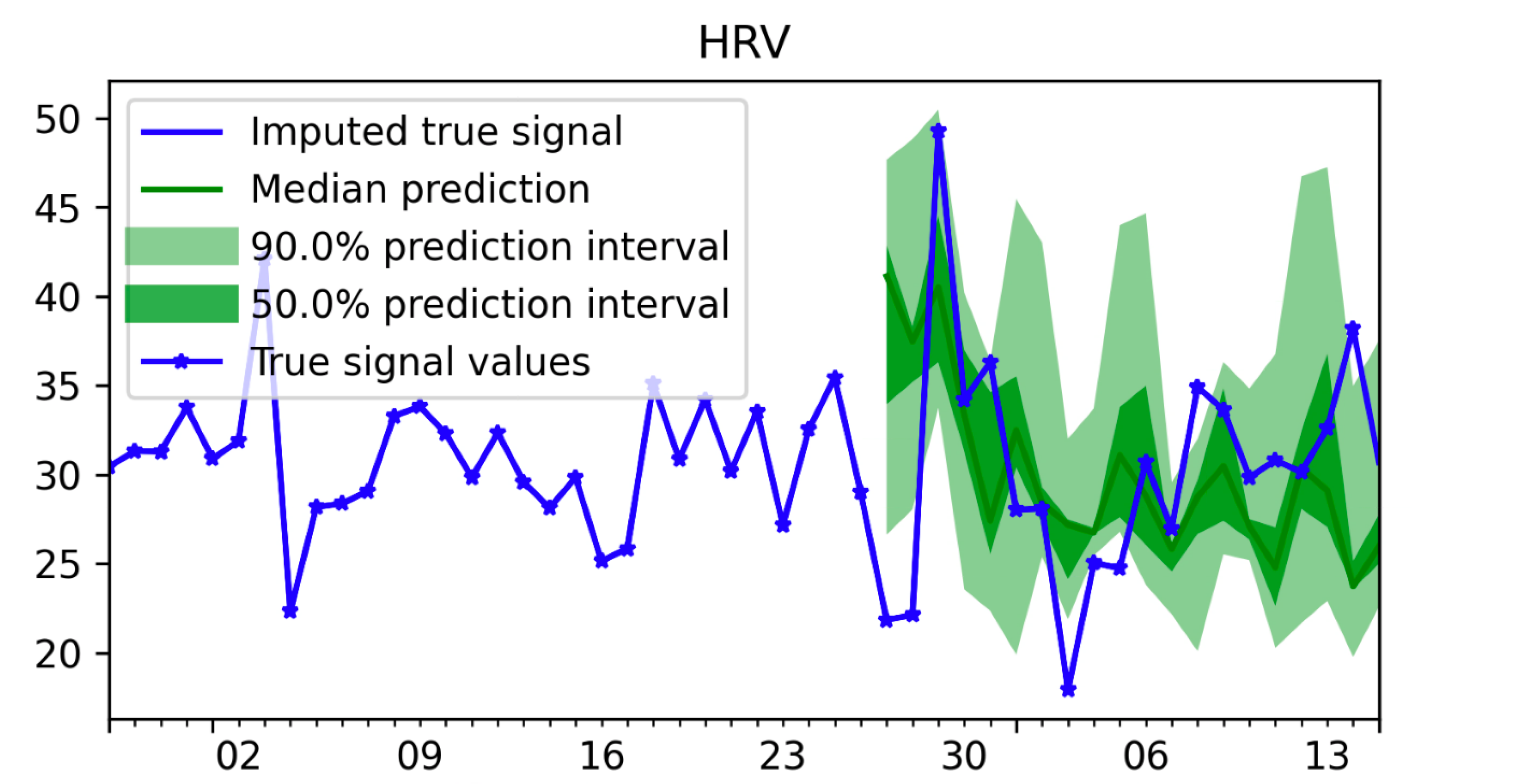}  & 
\includegraphics[width=.25\textwidth]{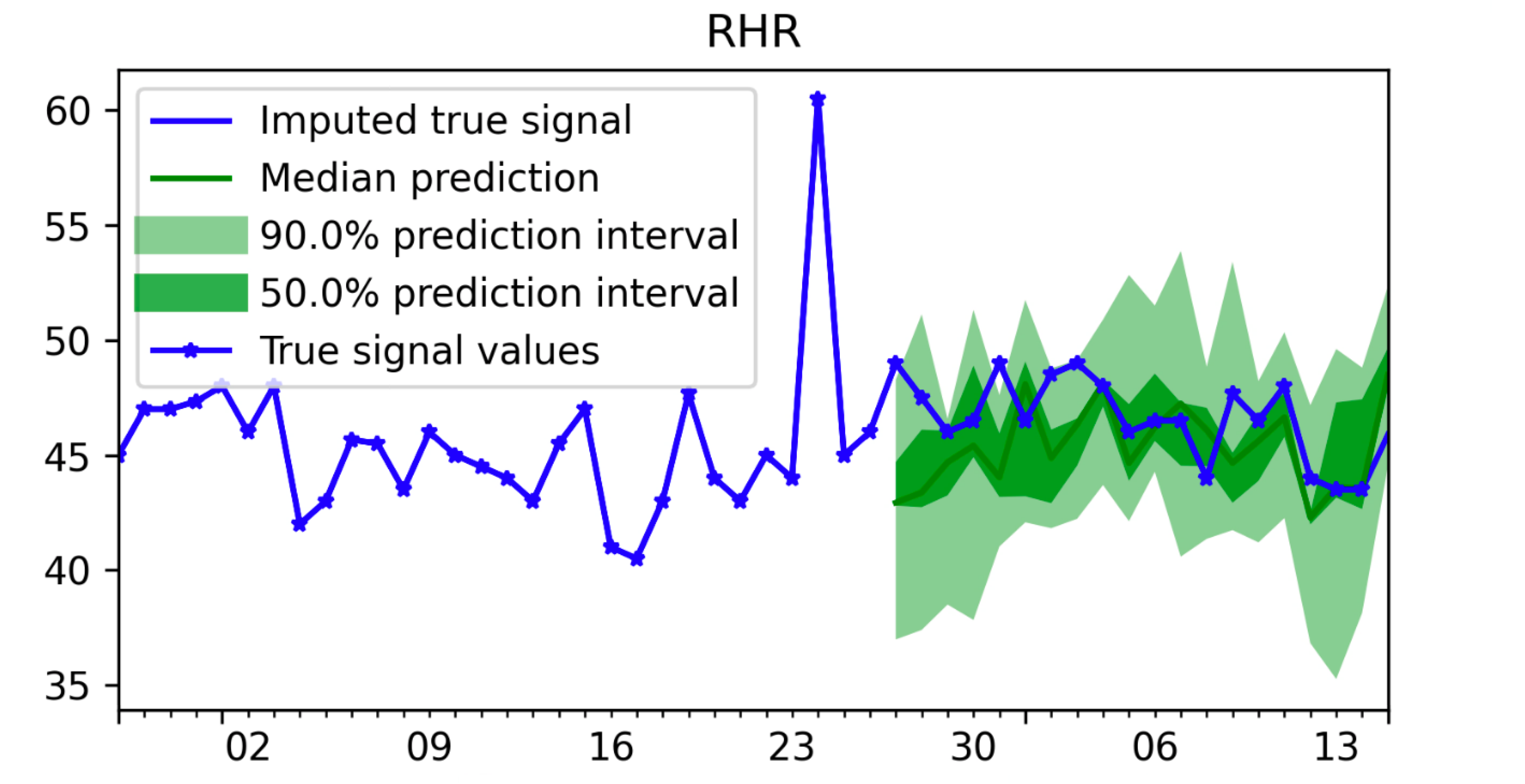}  & 
\includegraphics[width=.25\textwidth]{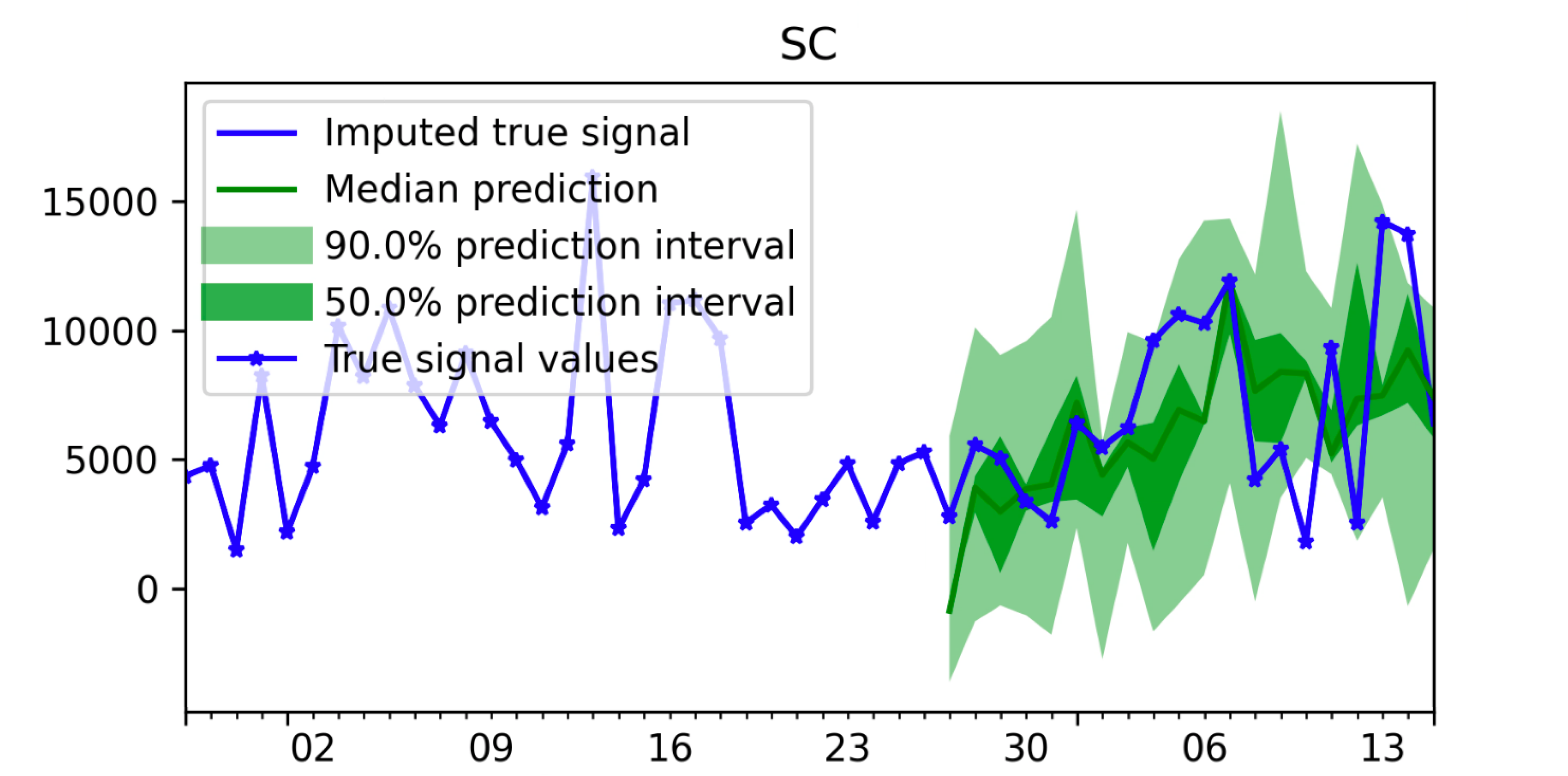}  & \\
\includegraphics[width=.25\textwidth]{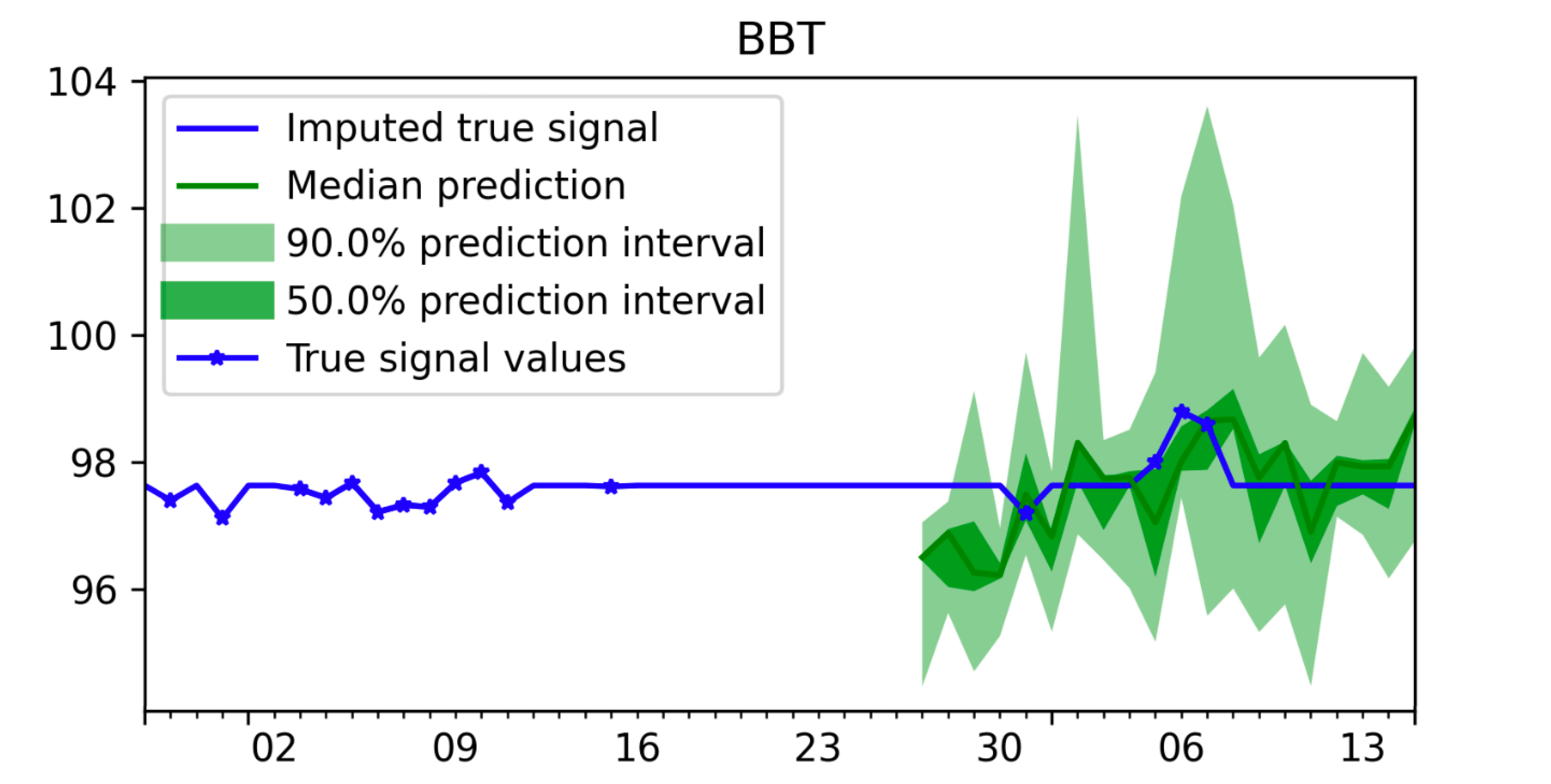}  & 
\includegraphics[width=.25\textwidth]{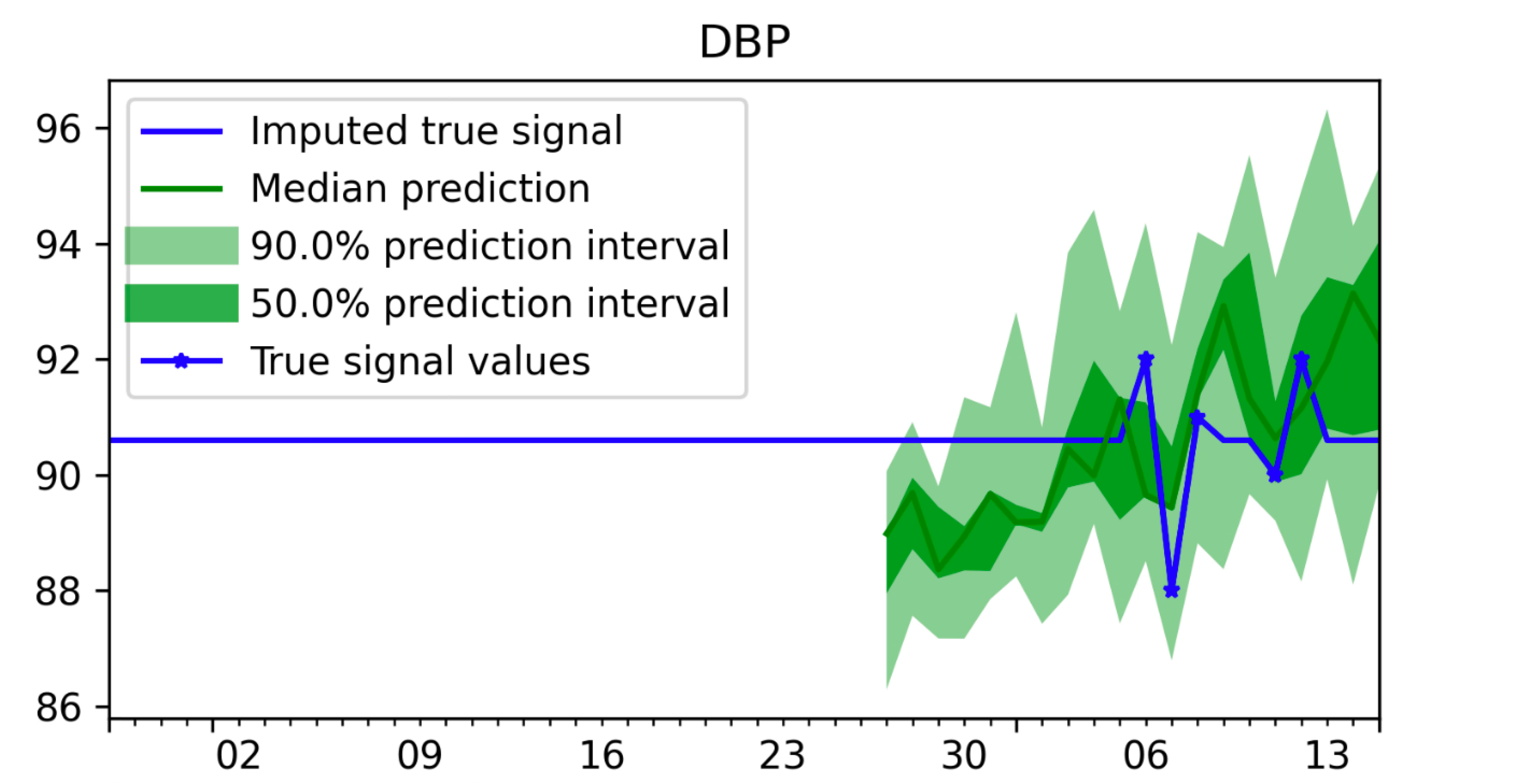}  & 
\includegraphics[width=.25\textwidth]{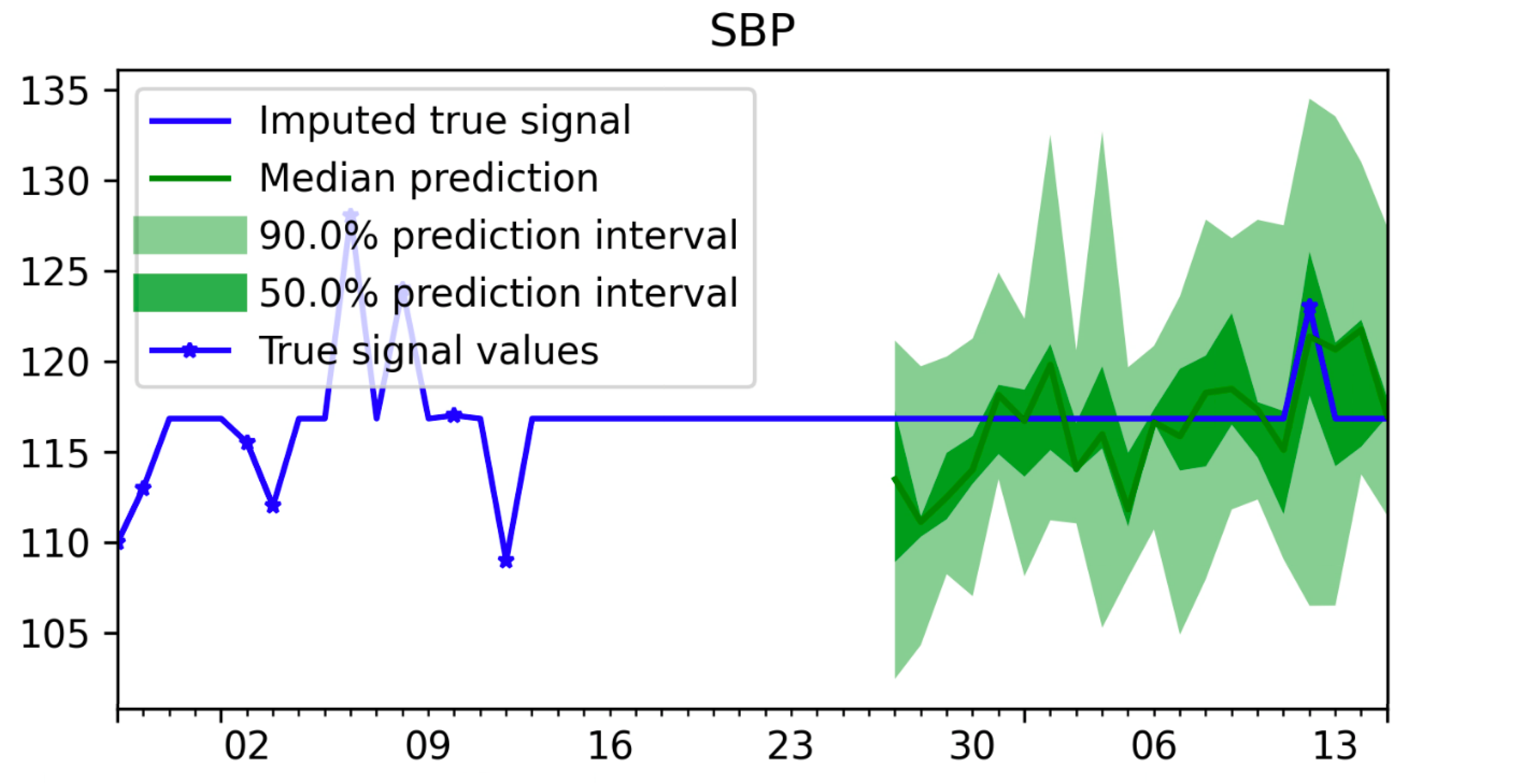}  & 
\\
\Xhline{2\arrayrulewidth}
            \includegraphics[width=.25\textwidth]{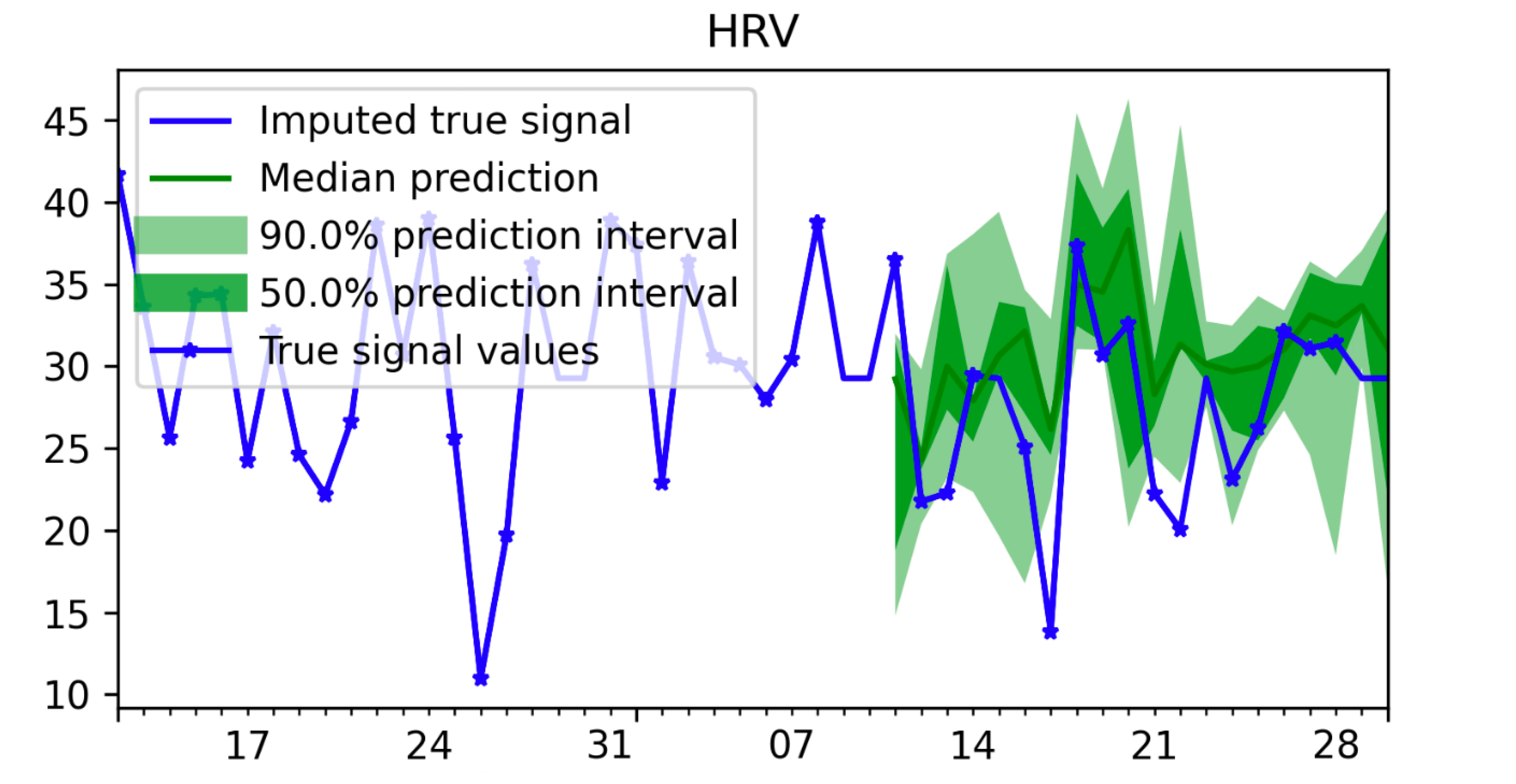}  & 
\includegraphics[width=.25\textwidth]{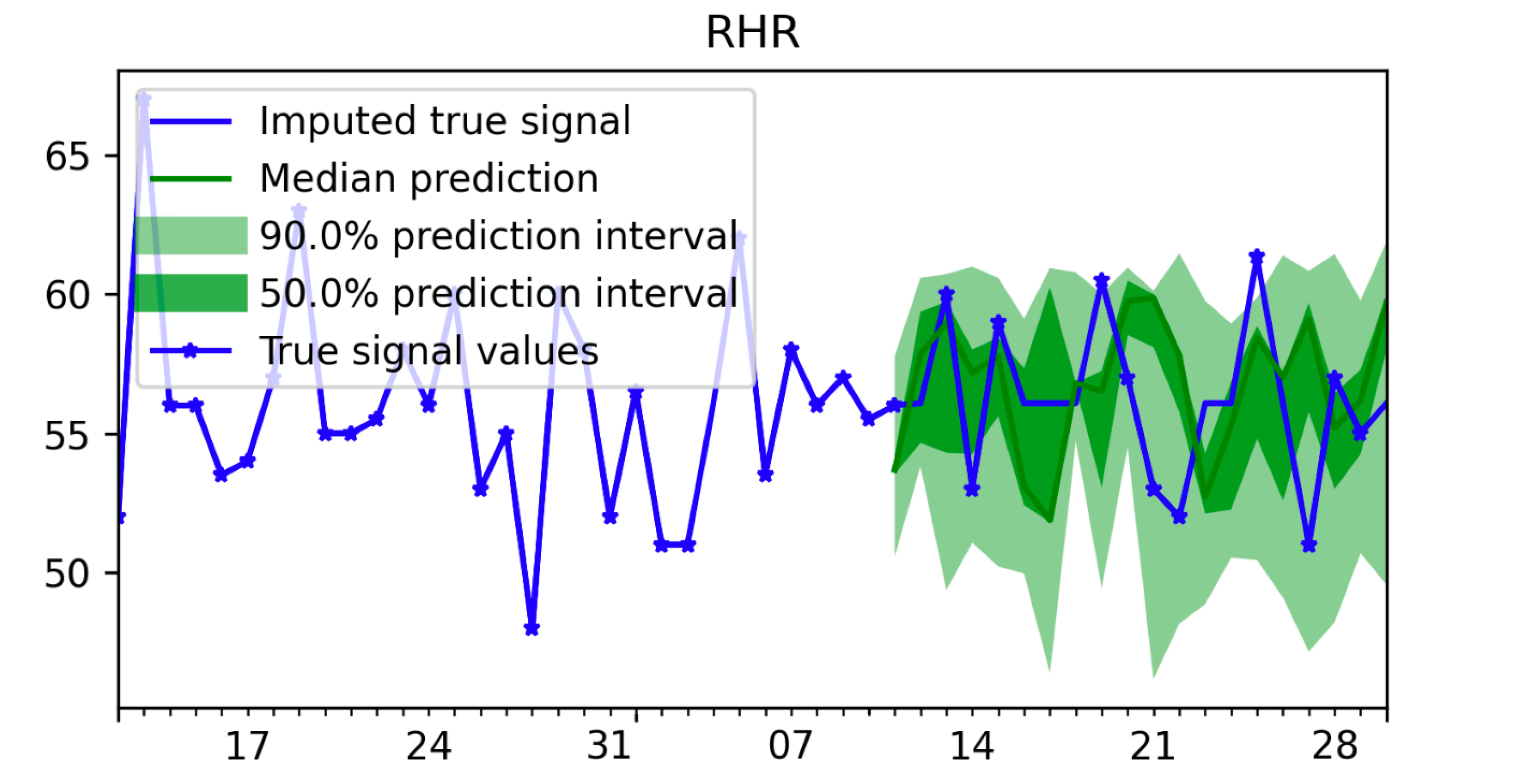}  & 
\includegraphics[width=.25\textwidth]{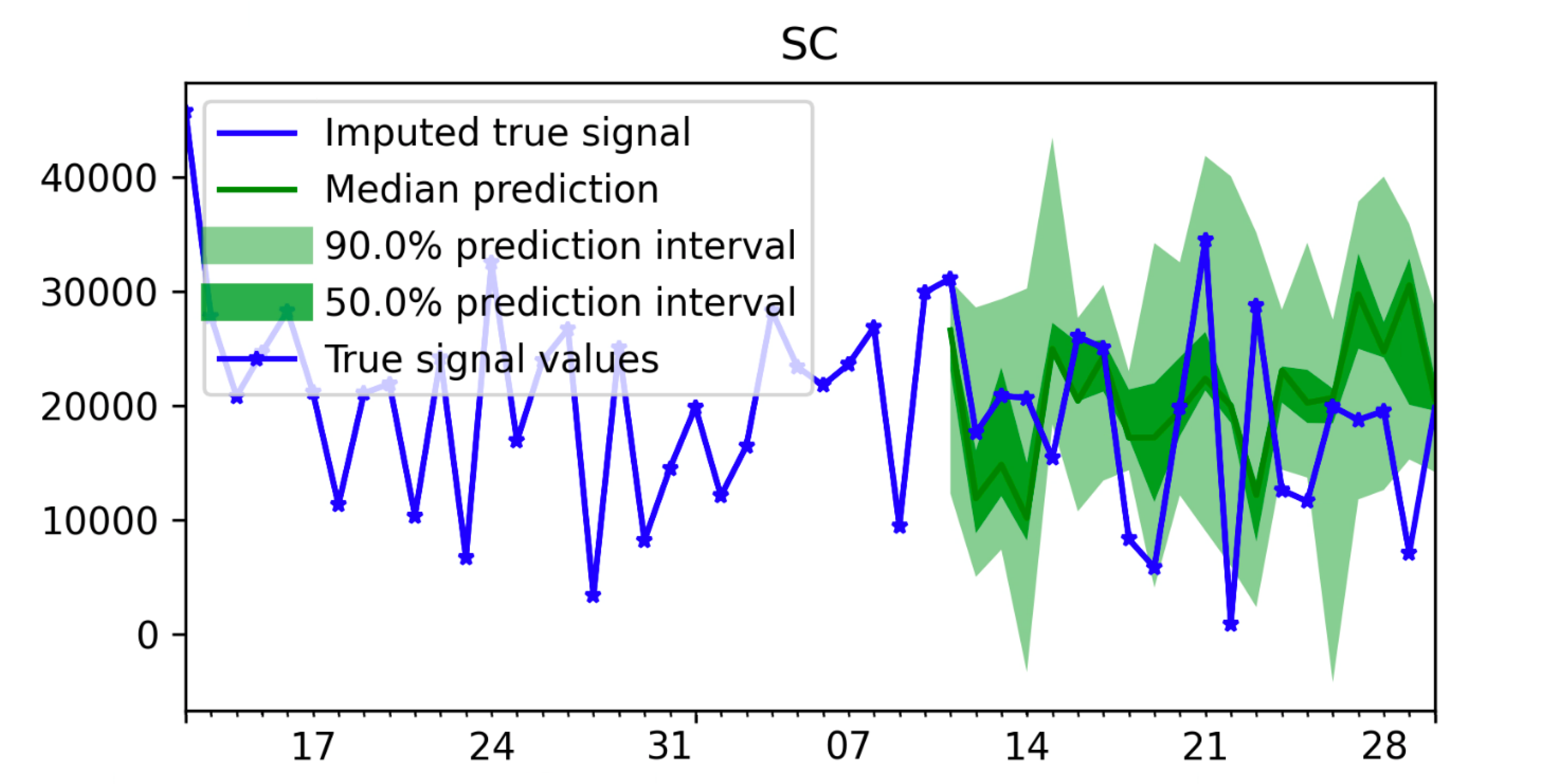}  & \\
\includegraphics[width=.25\textwidth]{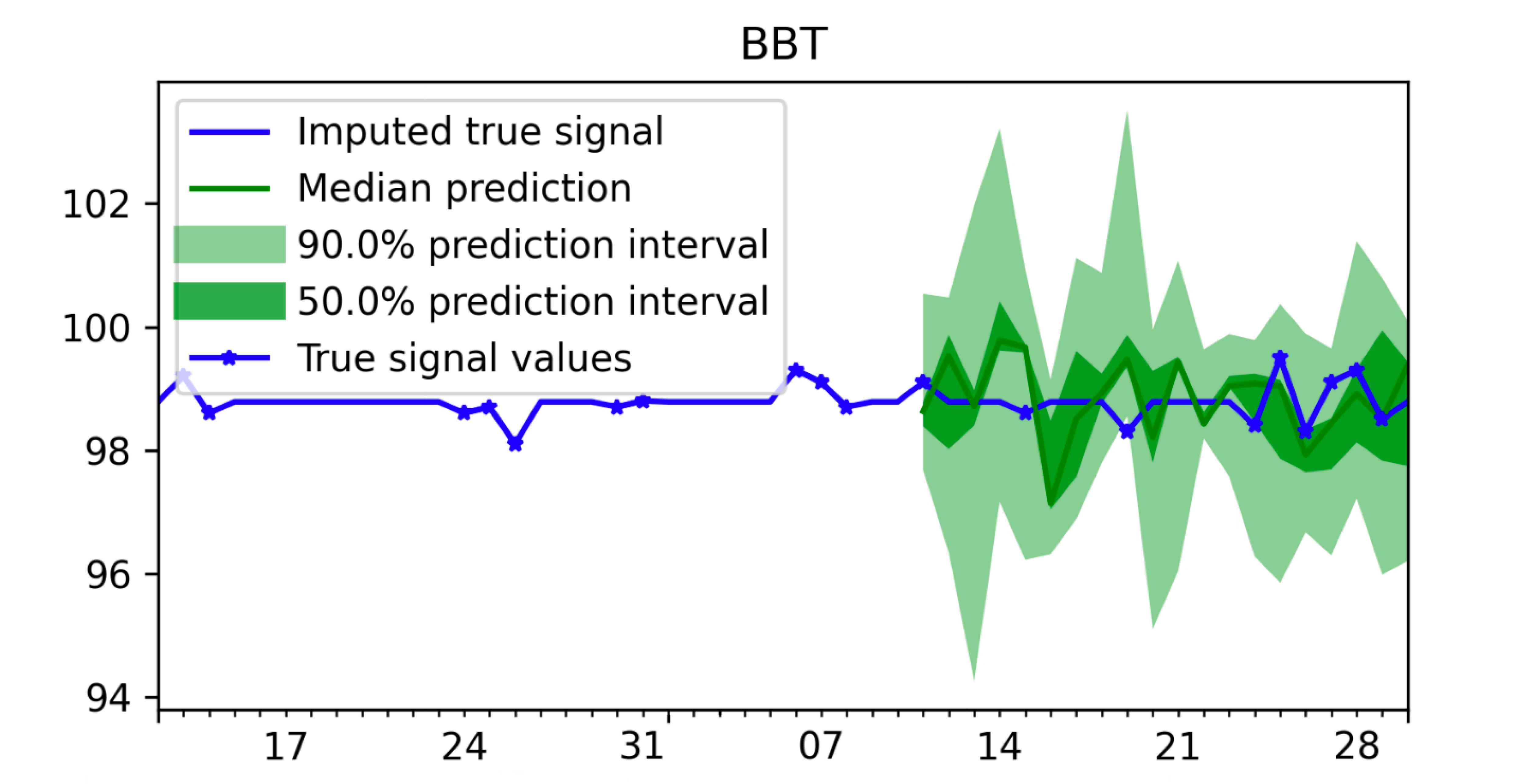}  & 
\includegraphics[width=.25\textwidth]{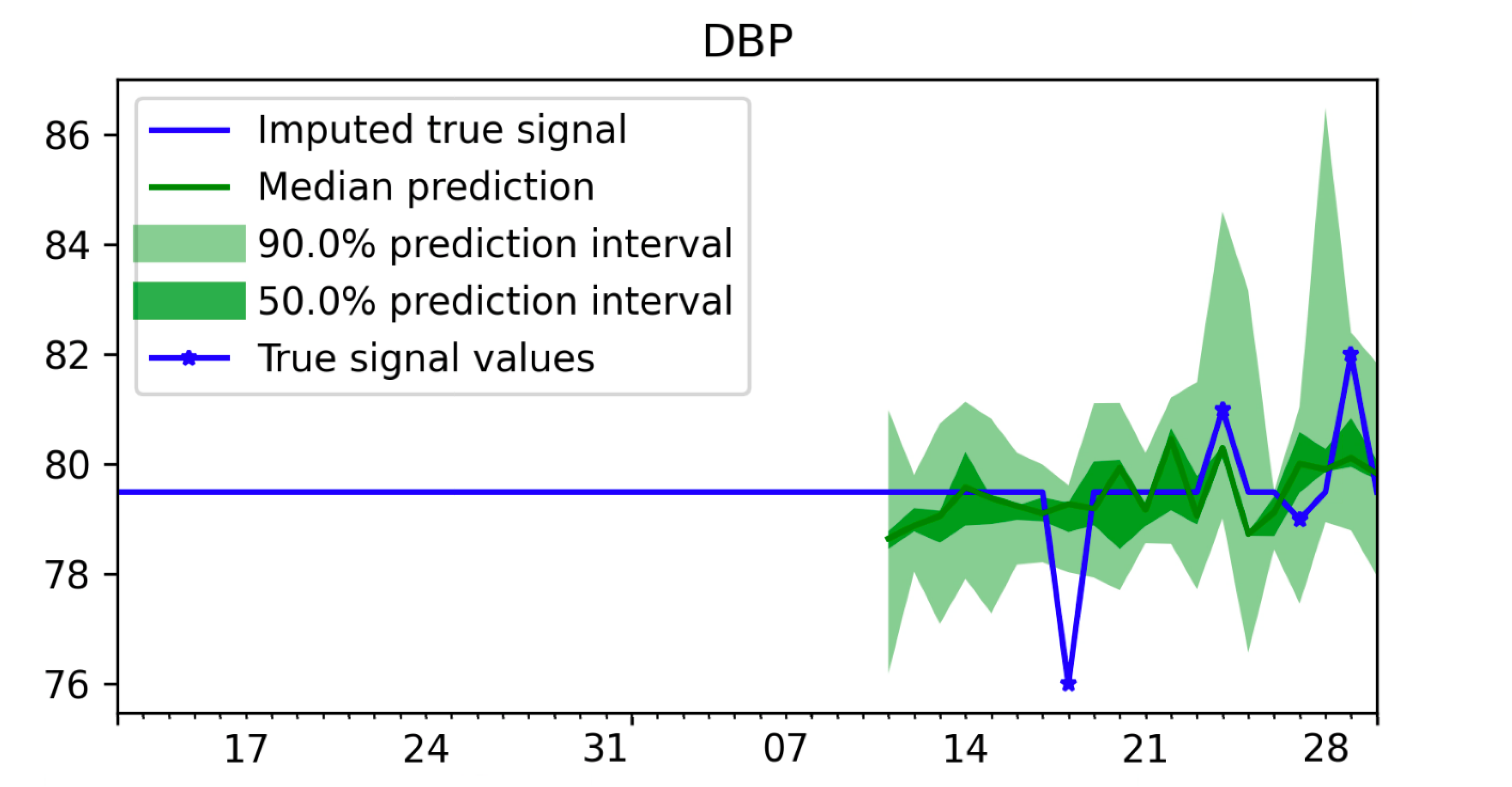}  & 
\includegraphics[width=.25\textwidth]{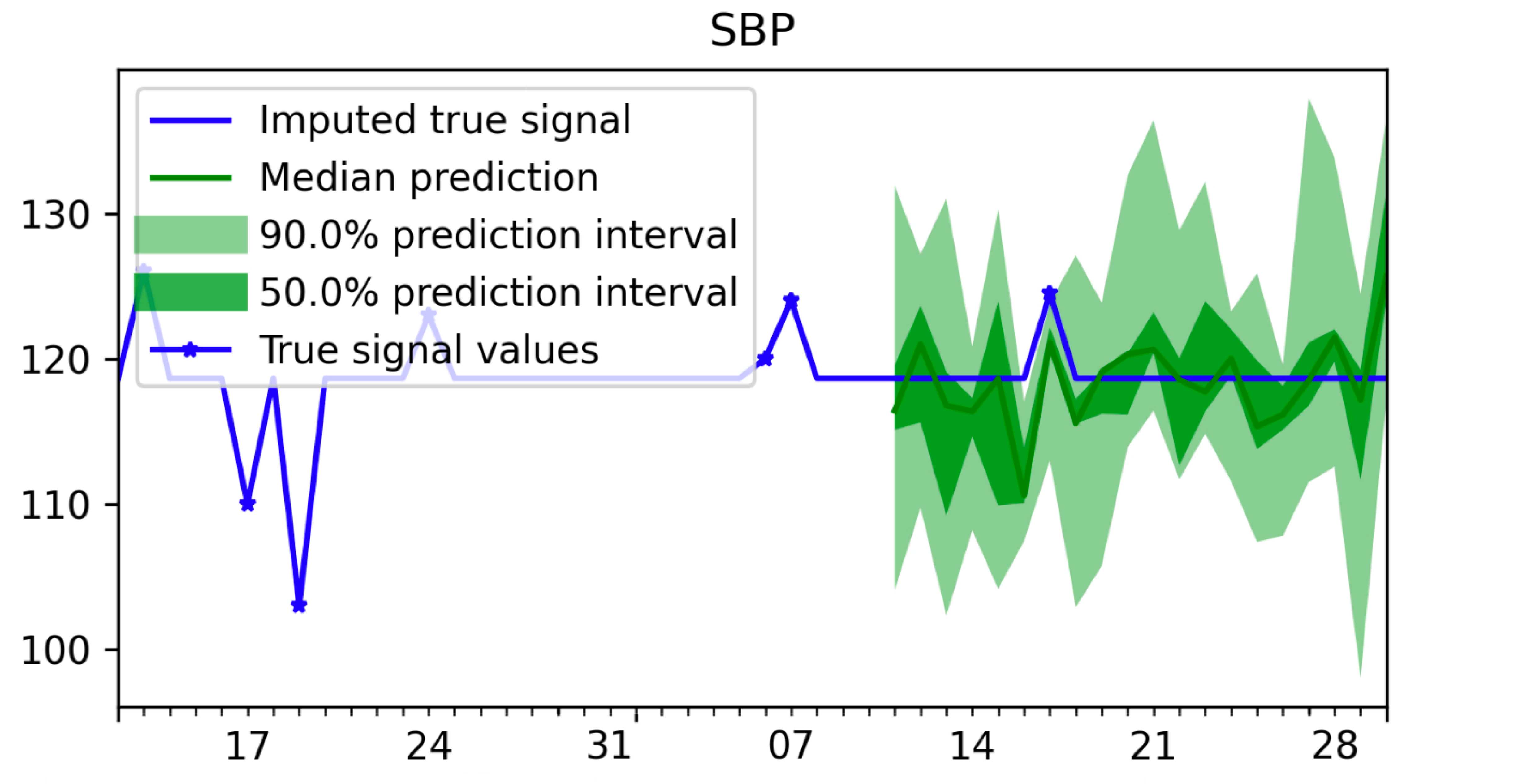}  & 
\\\Xhline{2\arrayrulewidth}
            \includegraphics[width=.25\textwidth]{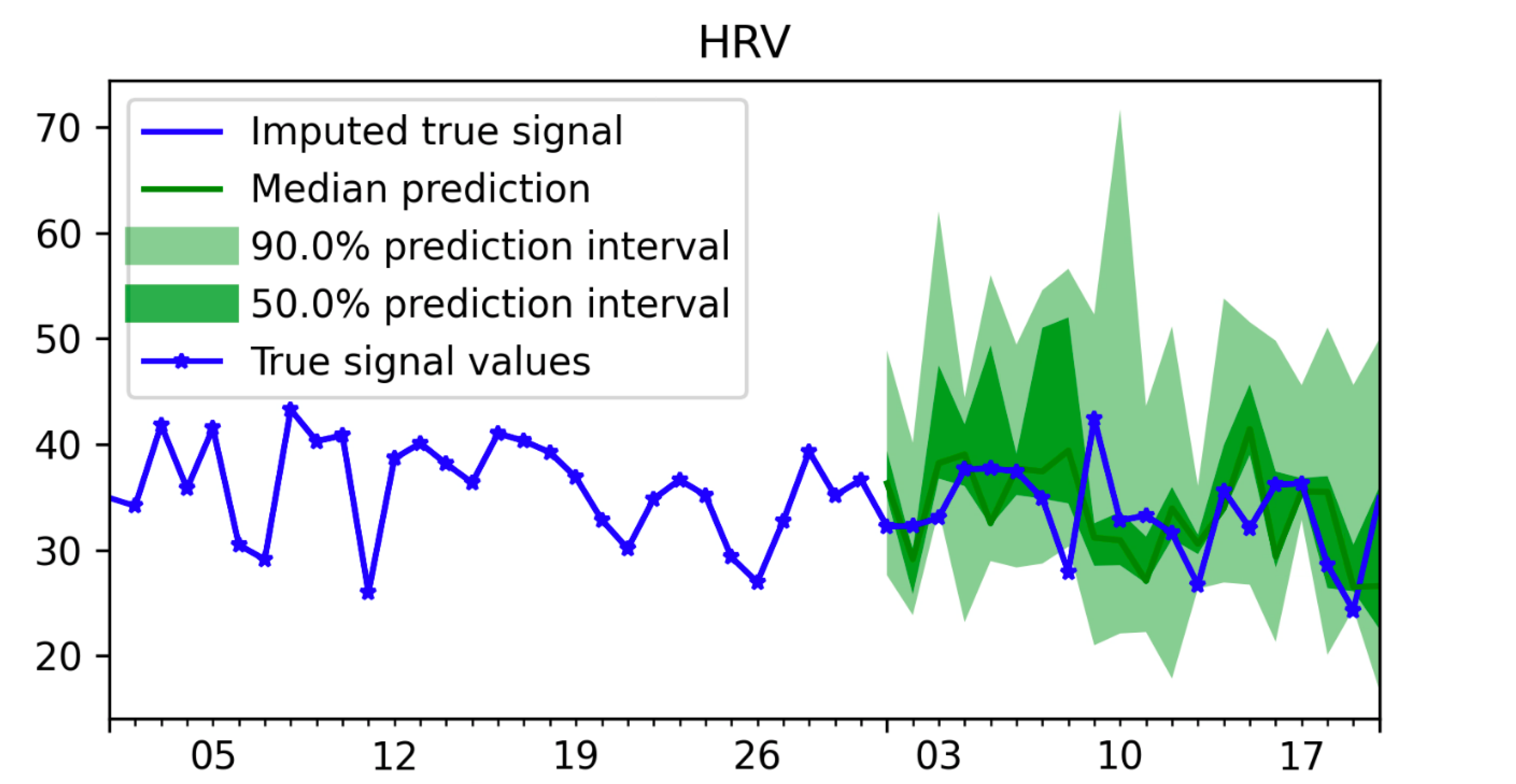}  & 
\includegraphics[width=.25\textwidth]{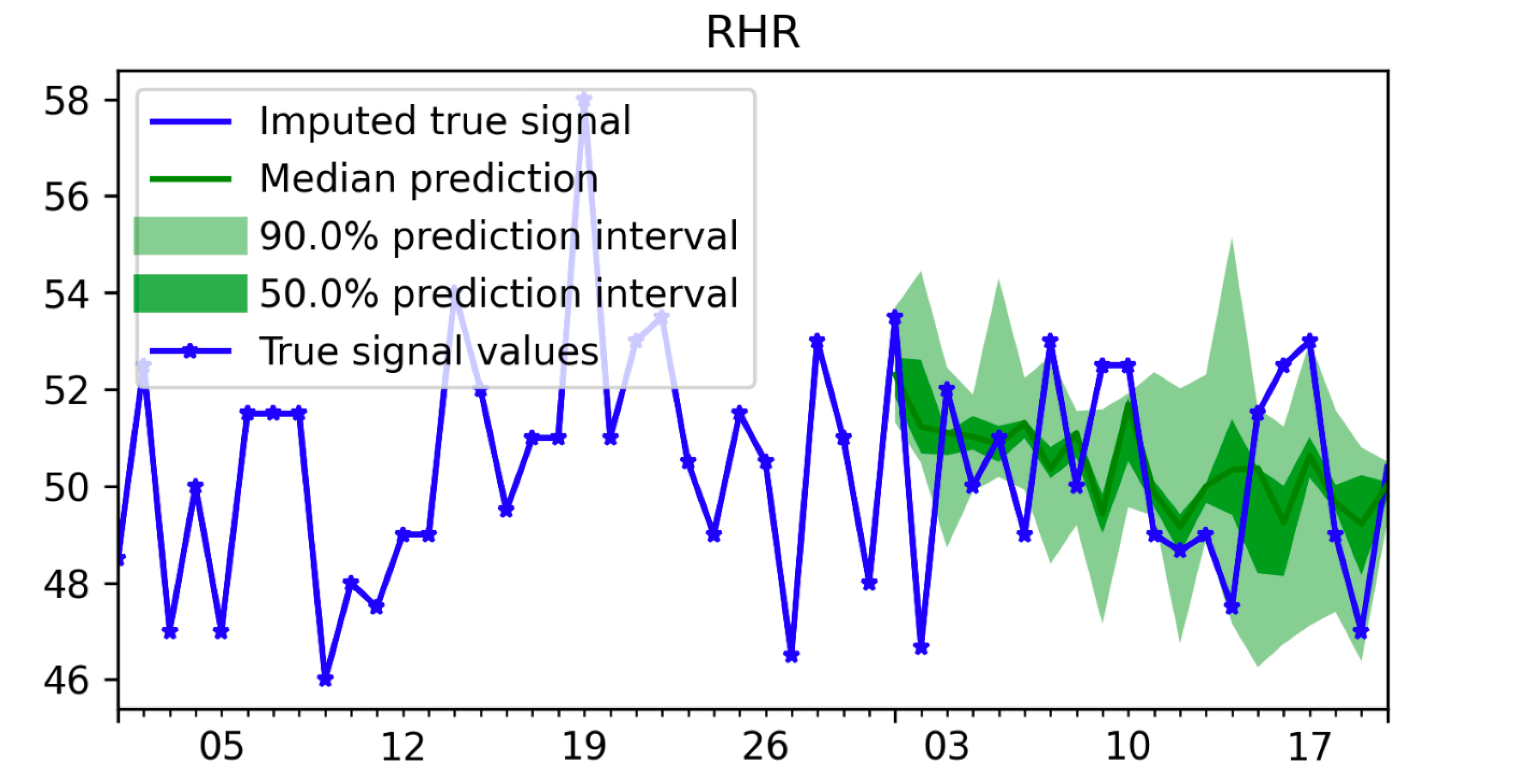}  & 
\includegraphics[width=.25\textwidth]{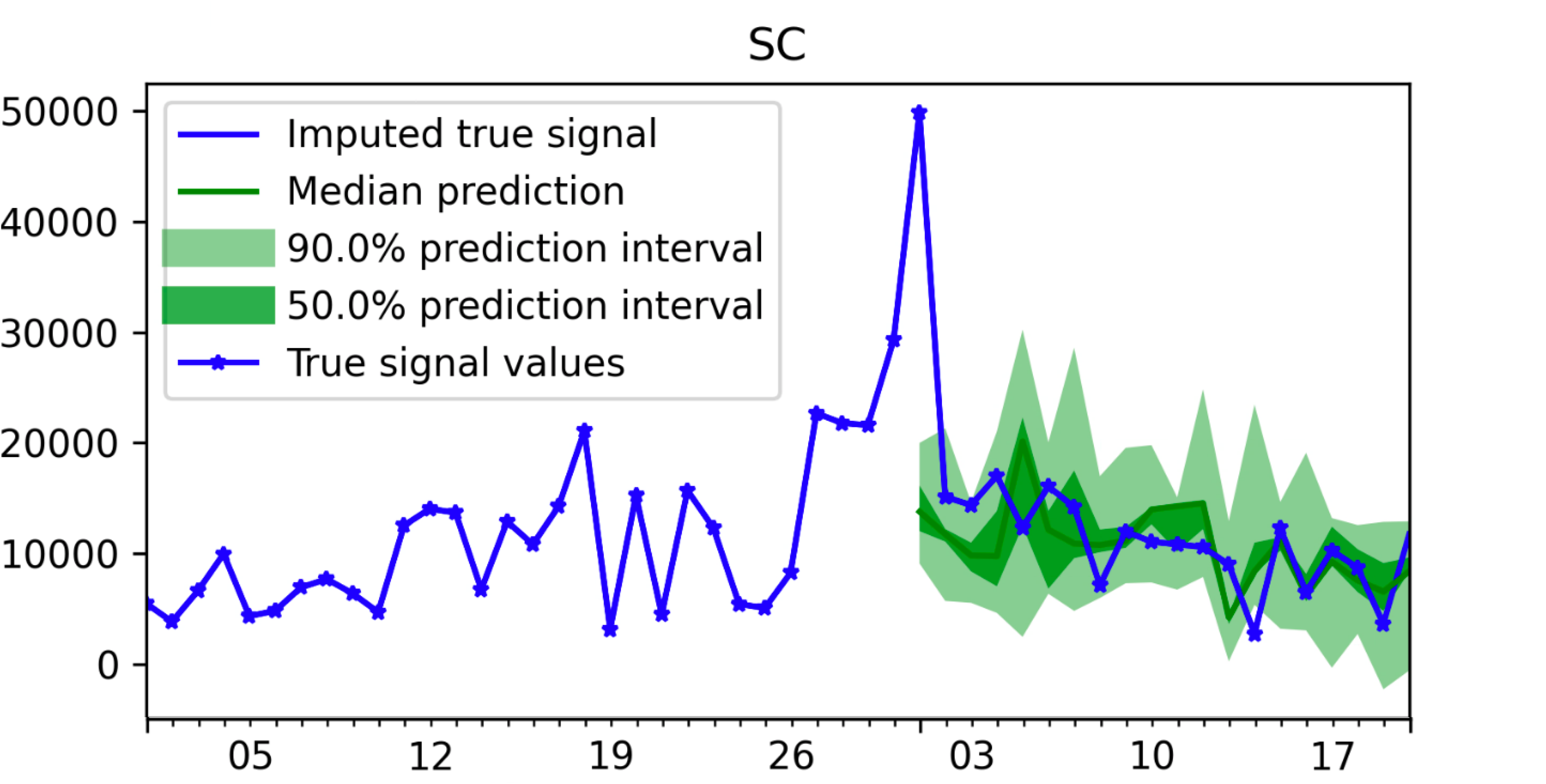}  & \\
\includegraphics[width=.25\textwidth]{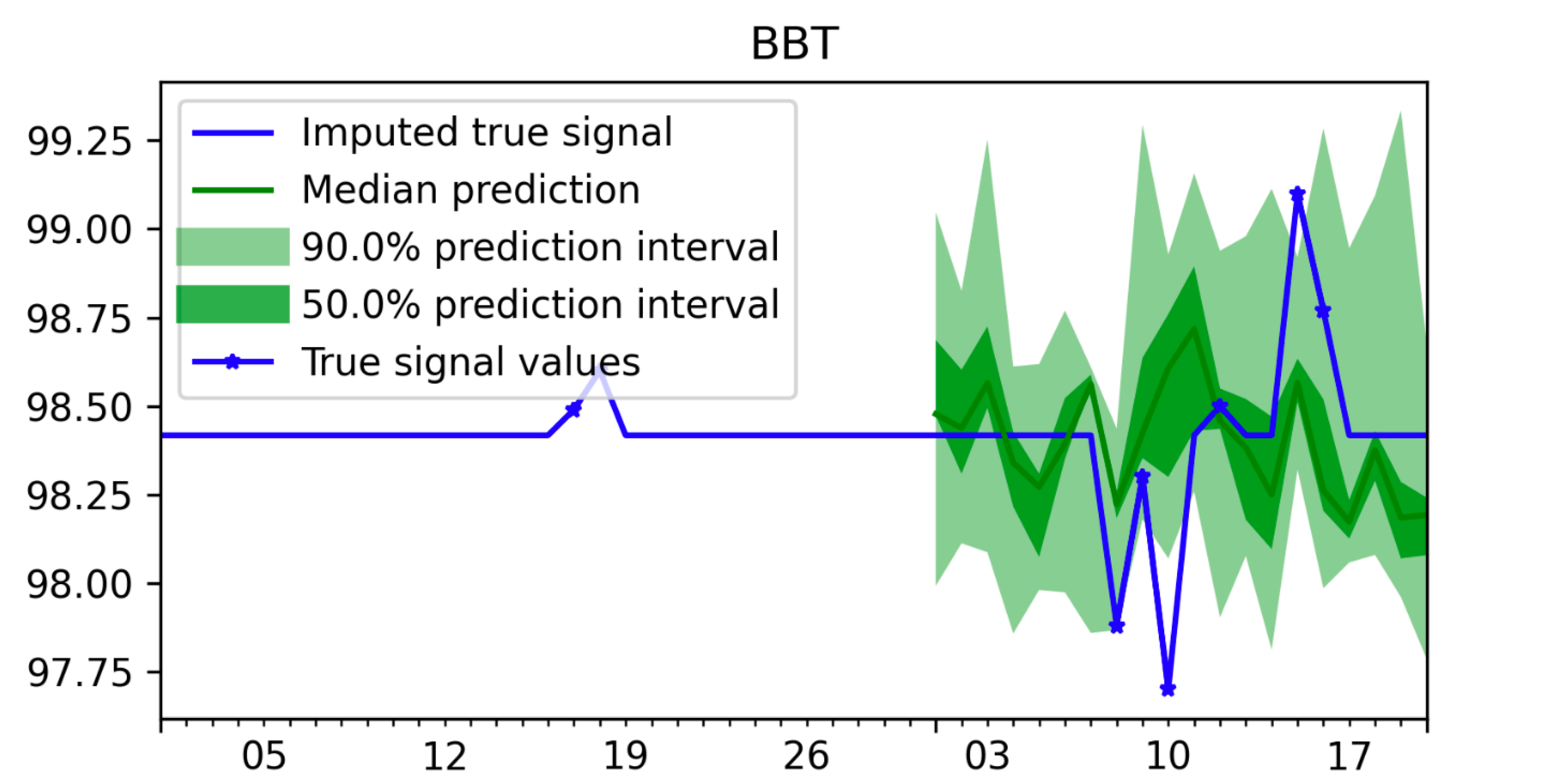}  & 
\includegraphics[width=.25\textwidth]{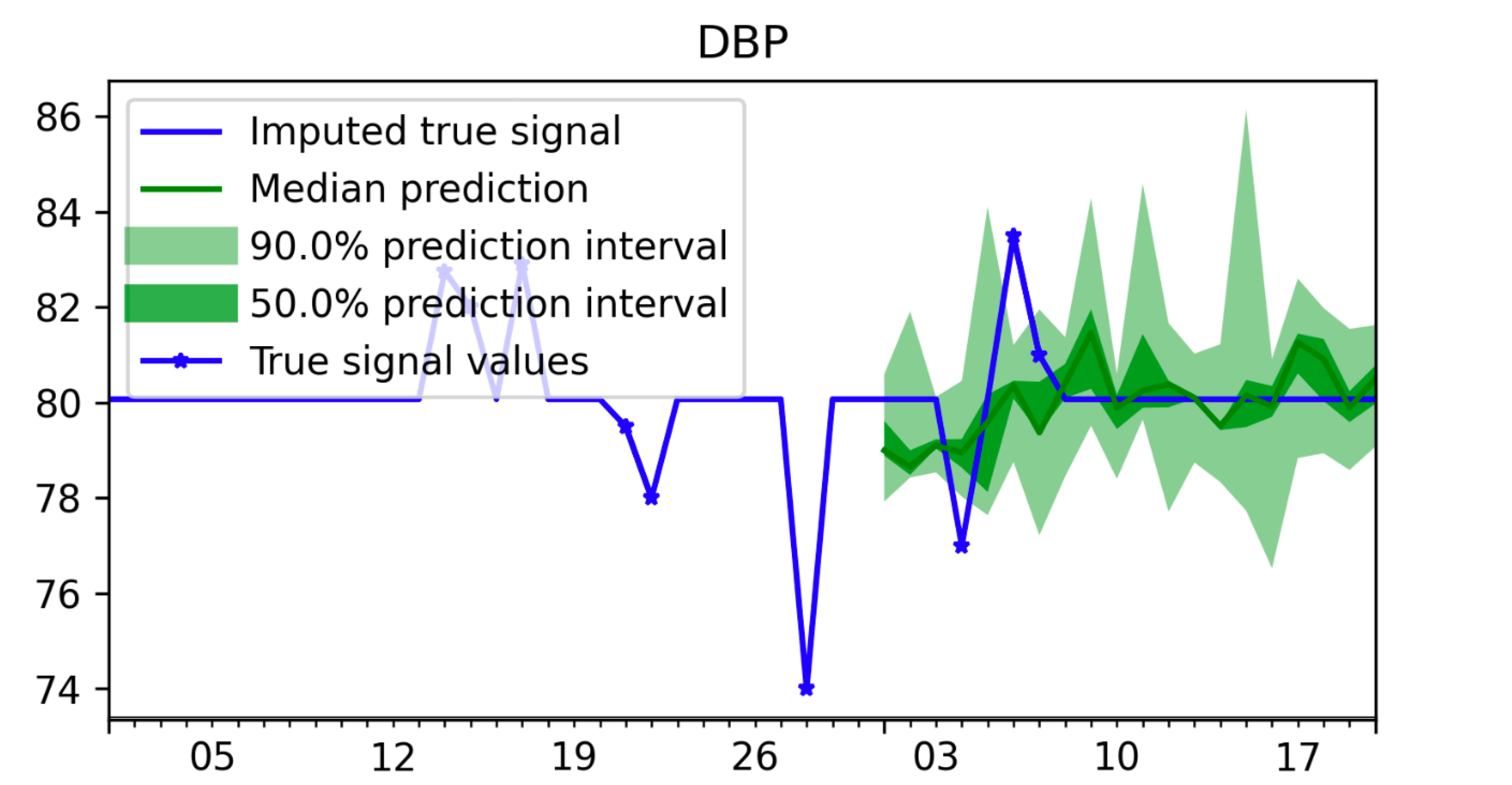}  & 
\includegraphics[width=.25\textwidth]{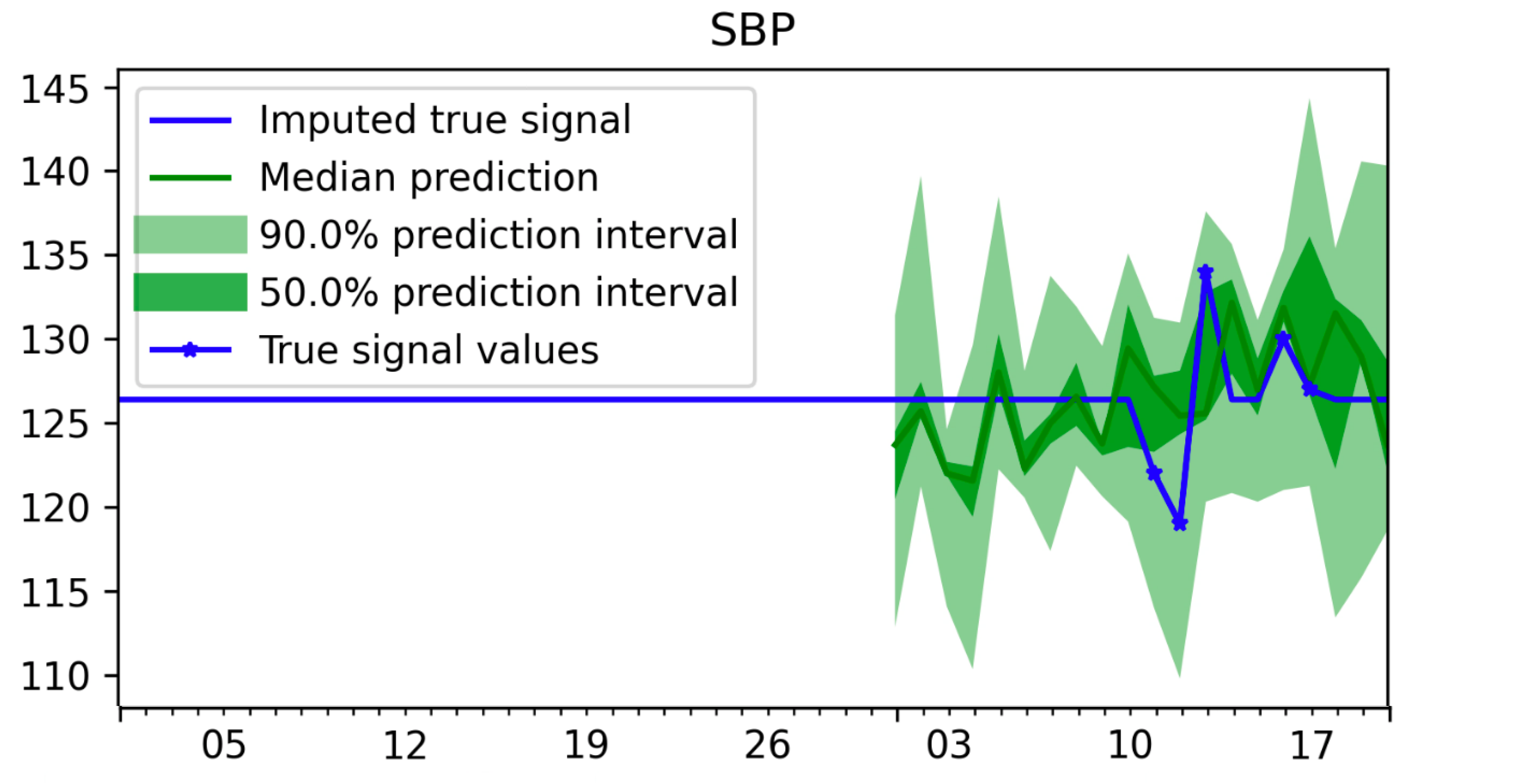}  & 
\\\Xhline{2\arrayrulewidth}
            \includegraphics[width=.25\textwidth]{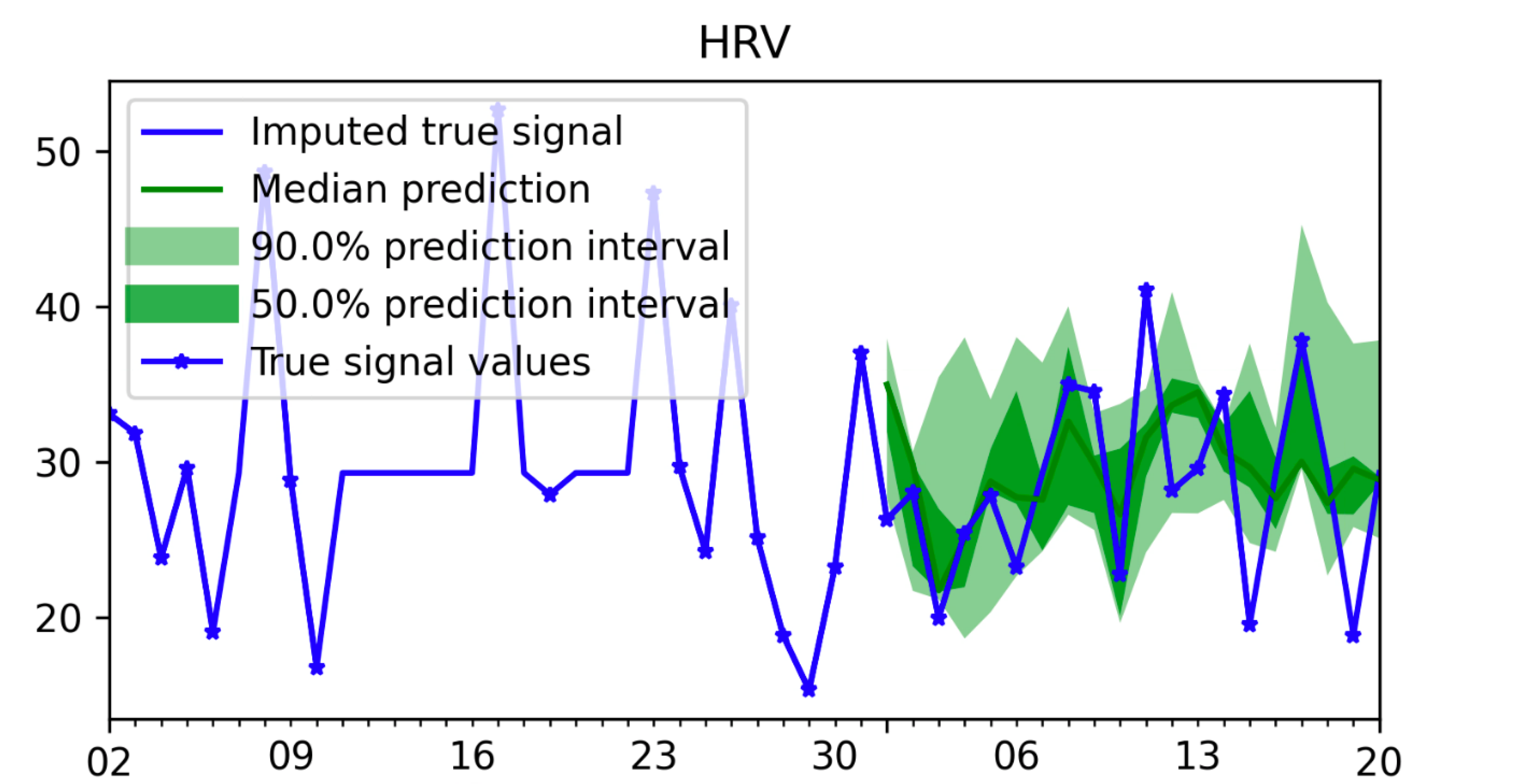}  & 
\includegraphics[width=.25\textwidth]{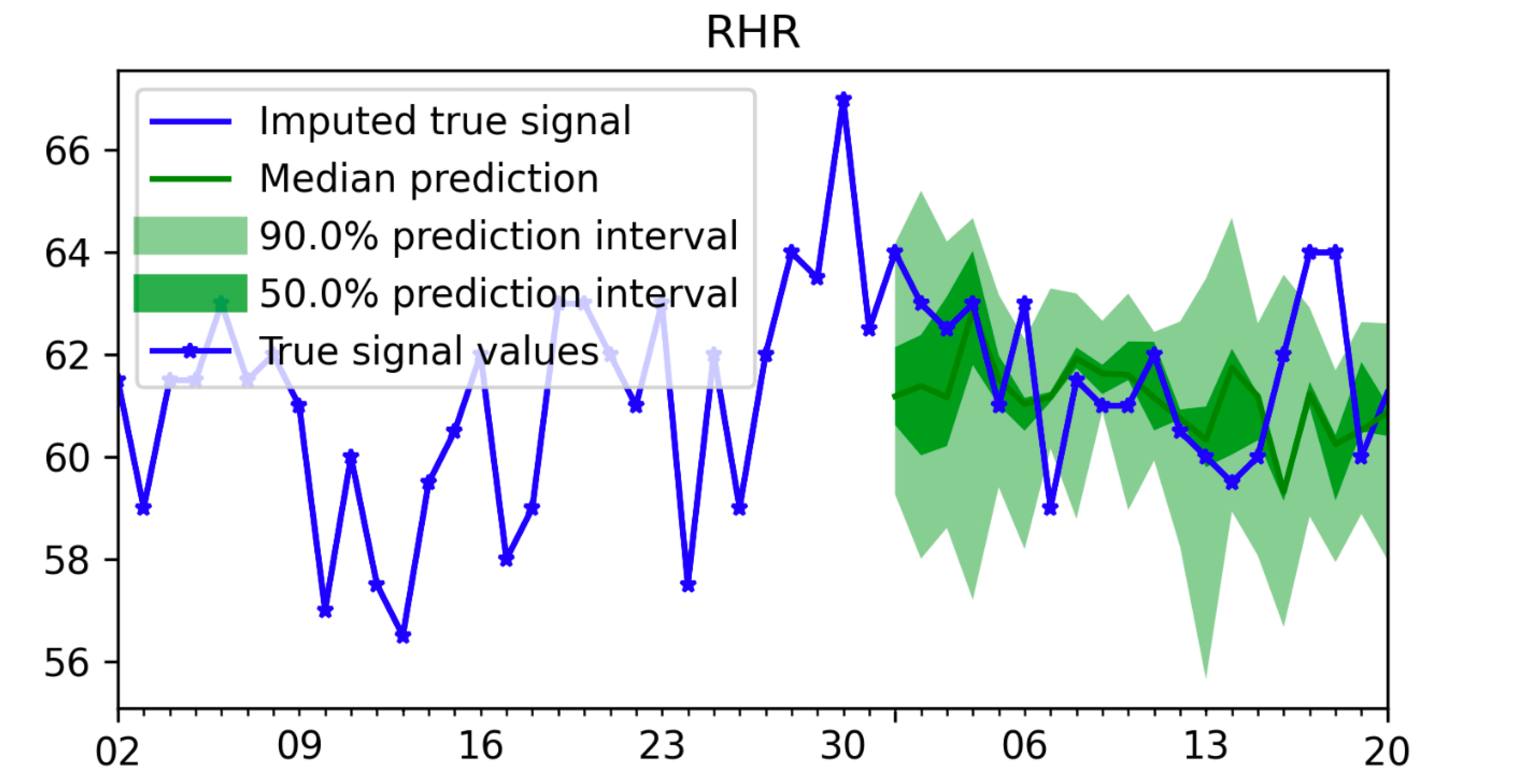}  & 
\includegraphics[width=.25\textwidth]{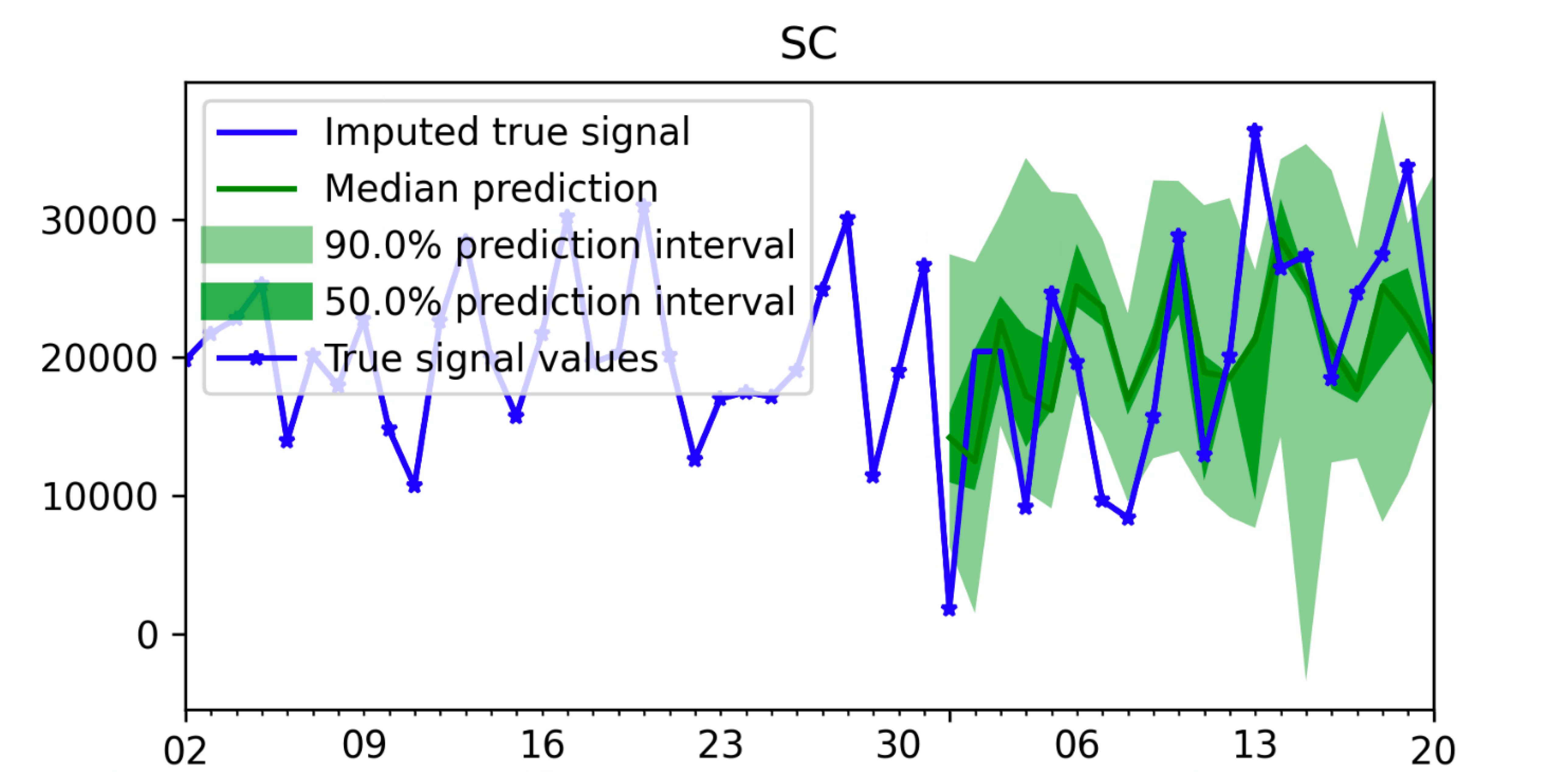}  & \\
\includegraphics[width=.25\textwidth]{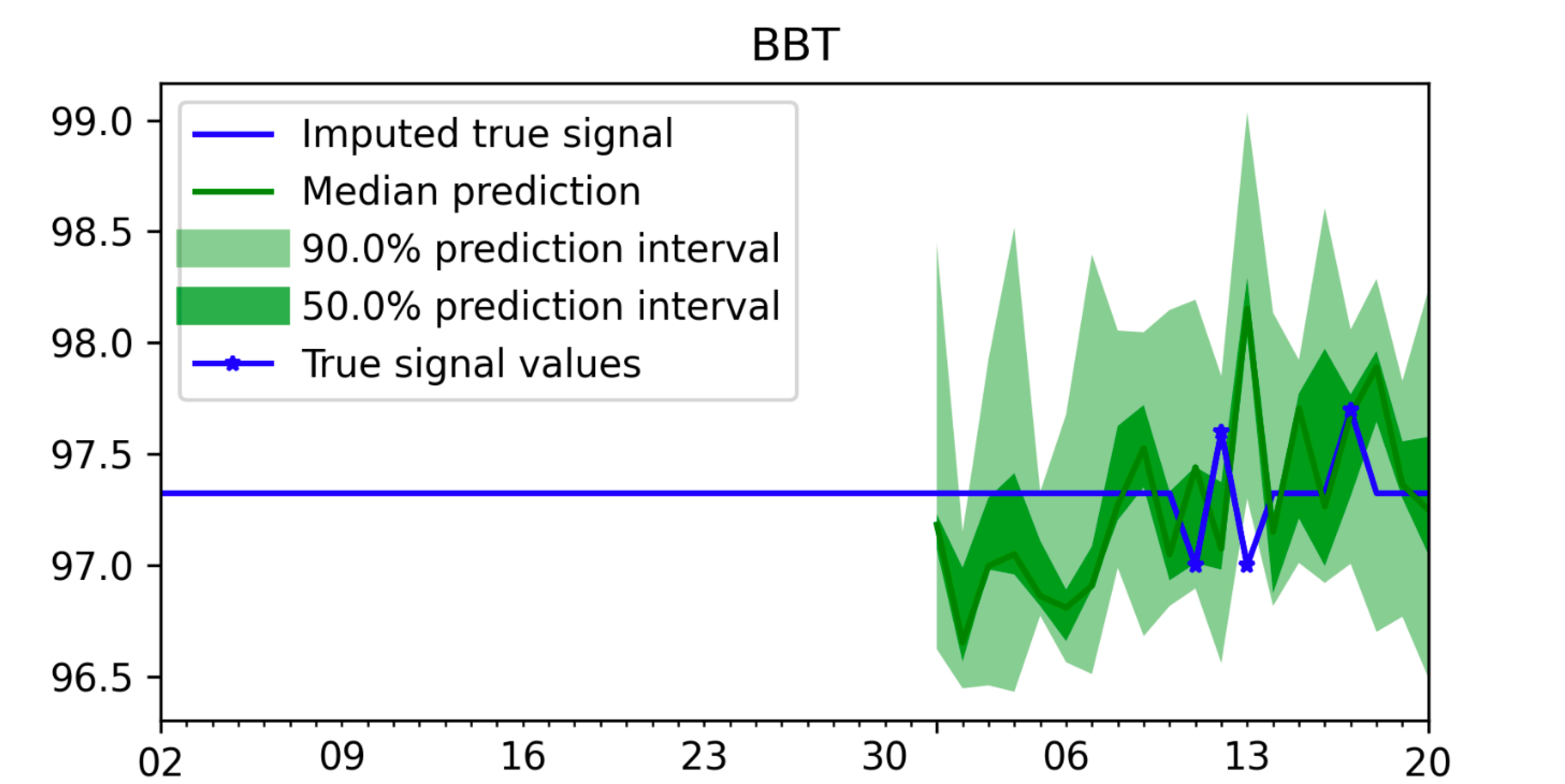}  & 
\includegraphics[width=.25\textwidth]{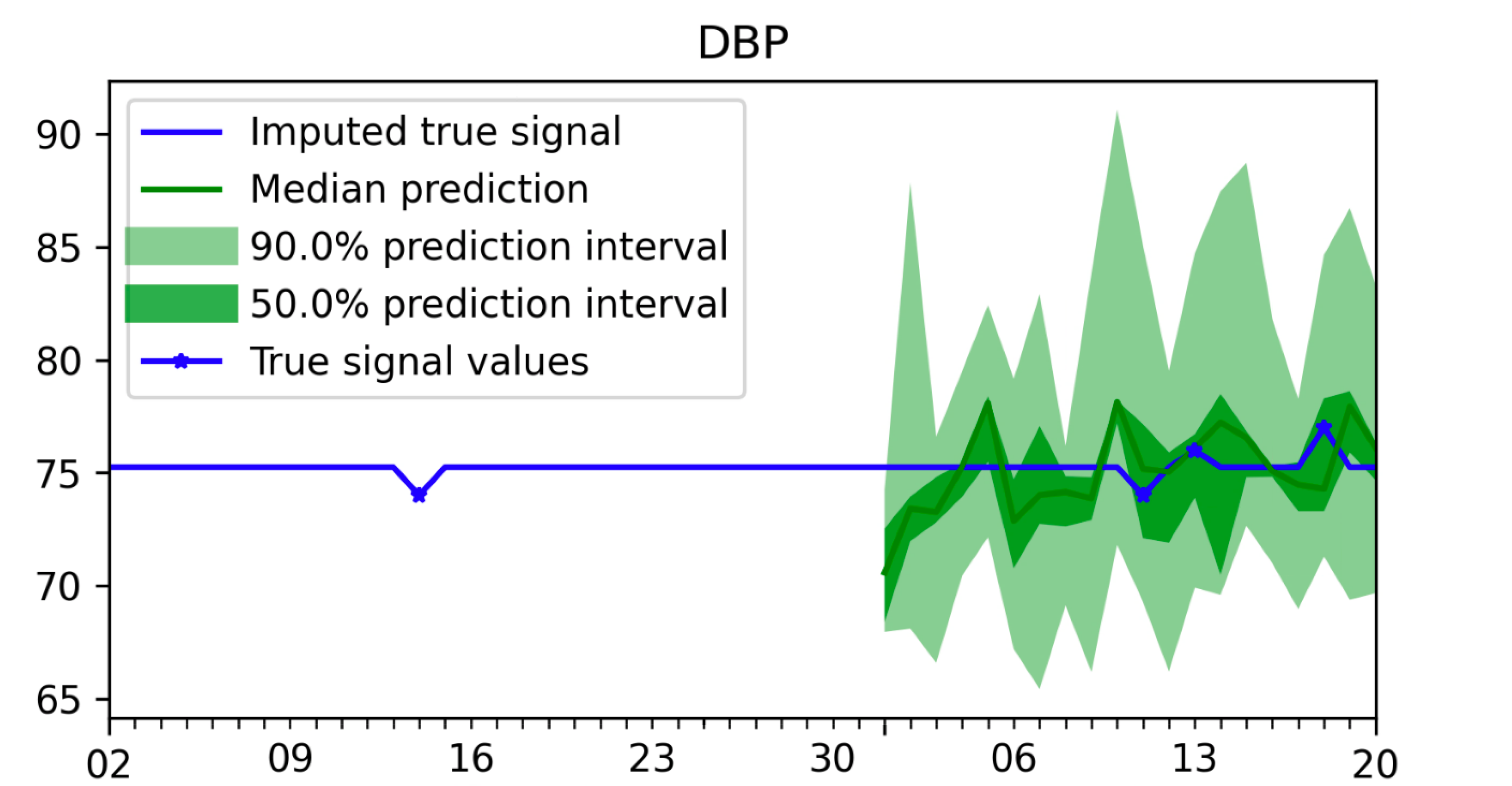}  & 
\includegraphics[width=.25\textwidth]{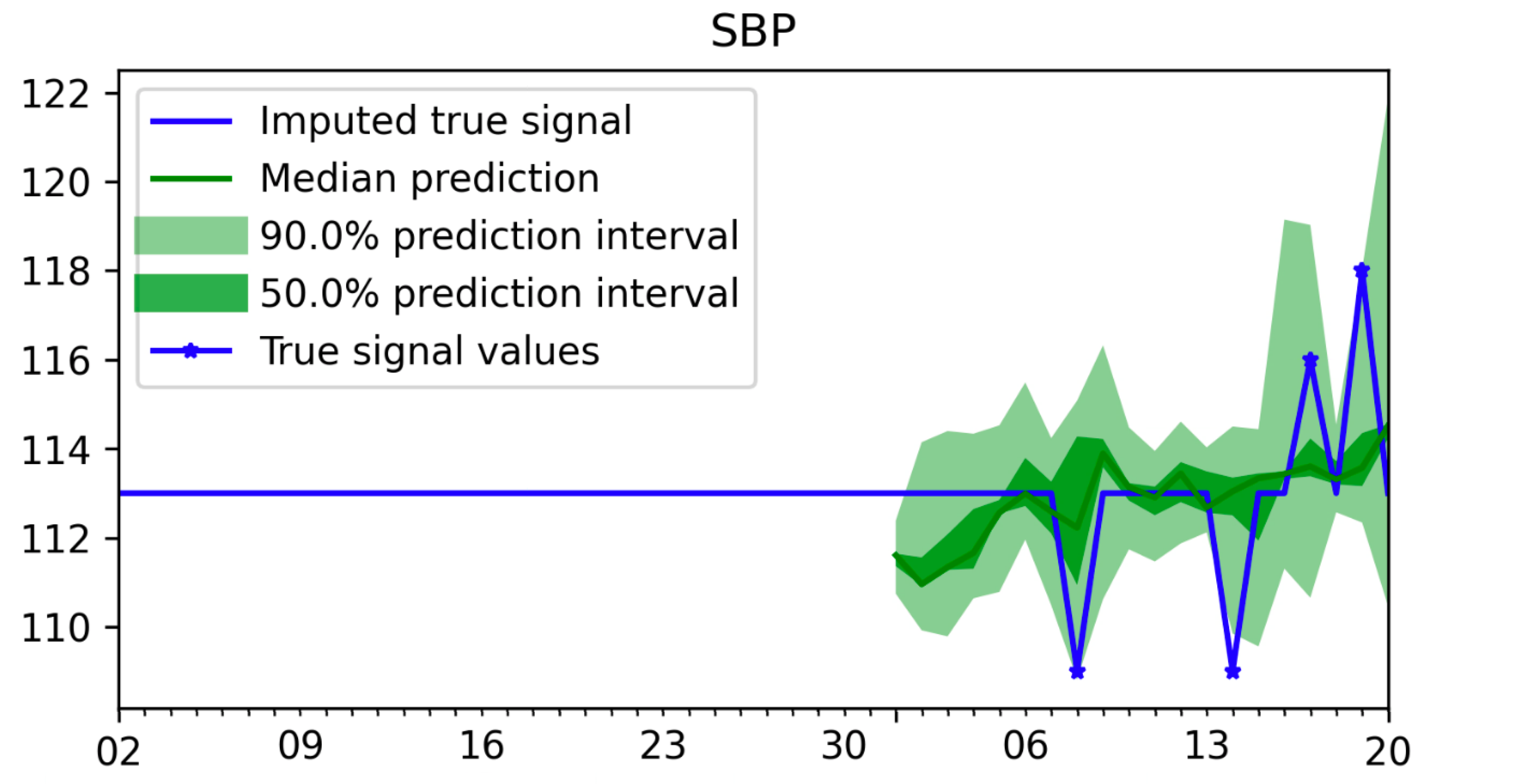}  & \\
\Xhline{2\arrayrulewidth}
      \end{tabular}
  \caption{Predicted signals and test set ground-truth for four additional  randomly selected subjects in the study (individual subjects are separated by horizontal lines).}
   \label{random4}
\end{center}  
\end{figure}

In Figure~\ref{maples_main}, we demonstrate the quality of predictions produced by the proposed model---we show the predicted median, $50\%$ and $90\%$ distribution intervals of for HRV, RHR, SC, BBT, DBP, and SBP in the future $20$ day window over a randomly chosen individual. In this case, the reduced dimensionality is $4$. To showcase the ability of the proposed model to generate precise forecasts (but also the dynamic complexity of signals -- it is well known that biomedical signals, such as HRV arise from complex nonlinear dynamical systems), more results of randomly selected individuals are presented in Figure~\ref{random4}. 

Although there have been many studies focusing on modeling sequential data, our method is especially effective for temporal bio-signals. In particular, modeling physiological signals is a complex task mainly because of (the potentially large number of) highly-correlated multivariate time-series, which reflect complex physiological interactions. Thus, we need a forecasting model that can account for the dependent high-dimensional nature of the underlying sensor signals, and at the same time a model that allows data to ``speak for themselves'' (i.e., without restrictive assumptions on the distribution class of each bio-signal). Additionally, models that assume inputs to be purely periodic signals, might capture rough trends in bio-signals -- periodic models fail to capture sharp changes in signals such as acute HRV depression due to sickness or a sharp increase in step counts due to a non-stationary behavioral or environmental change such as a change in schedule due to travel. Our model, on the other hand, explicitly learns the latent space structure through an expressive time-conditioned distribution model of temporal transitions without the strong assumptions of traditional multivariate (probabilistic) models. At the same time, its probabilistic nature is able to quantify the predictive uncertainty, which is advantageous in real-world critical applications.

Interpretable representation learning on time-series is a fundamental problem for uncovering the latent structure in complex systems, such as sensor streams of vital signals.
To reveal the underlying factors controlling accurate forecasts, we train the model with $D=2$ and $D=3$, and visualize the representations across 50 time points for eight different individuals.
These individuals belong to two disjoint cardiorespiratory fitness levels: the first group consists of four subjects with low values of estimated $\text{VO}_2\text{max}$~\citep{applevo2max}, $\text{VO}_2\text{max}<25$, and the second group consists of four subjects with high values of $\text{VO}_2 \text{max}, \text{VO}_2\text{max}>40$. Maximum oxygen consumption ($\text{VO}_2\text{max}$) is the measurement of the maximum amount of oxygen a person can utilize during intense exercise and is considered the gold standard for determining an individual's cardiorespiratory fitness. The latent representations for both $D=2$ and $D=3$ indicate that the mapping learned by the optimization leads to a space with easily visualized clusters that well-separates individuals according to $\text{VO}_2\text{max}$ (Figure~\ref{latent_reps_2d3d}). Our latent space estimated using RHR, HRV, BBT, DBP, SBP, and SC corroborates other recent methods that associate step count, sedentary time, and moderate-vigorous physical activity with $\text{VO}_2\text{max}$~\citep{nayor2021physical}. 
\subsection{Results on publicly available datasets}

We use publicly available datasets to evaluate the accuracy of forecasts of future observations $\mathbf{y}_t \in \mathbf{R}^N$  compared to to a wide range of existing baselines (described in the Appendix). The \textbf{Solar} dataset, with $N=137$ and $T_\text{total}=4832$, contains hourly solar power production records. The \textbf{Electricity} dataset with $N=370$ and $T_\text{total}=32303$, is a hourly time-series of the electricity consumption of $370$ customers. The \textbf{Traffic} dataset, with $N=963$ and $T_\text{total}=4049$, contains hourly occupancy rates of car lanes. \textbf{Wiki}, with $N=2000$ and $T_\text{total}=792$, is a daily time-series daily page views of Wikipedia articles. \textbf{Taxi}, with $N=1214$ and $T_\text{total}=1488$, is a spatio-temporal traffic time-series of New York taxi rides taken every $30$ minutes.

 \begin{figure}[H]
\begin{center}
        \includegraphics[width=.23\textwidth]{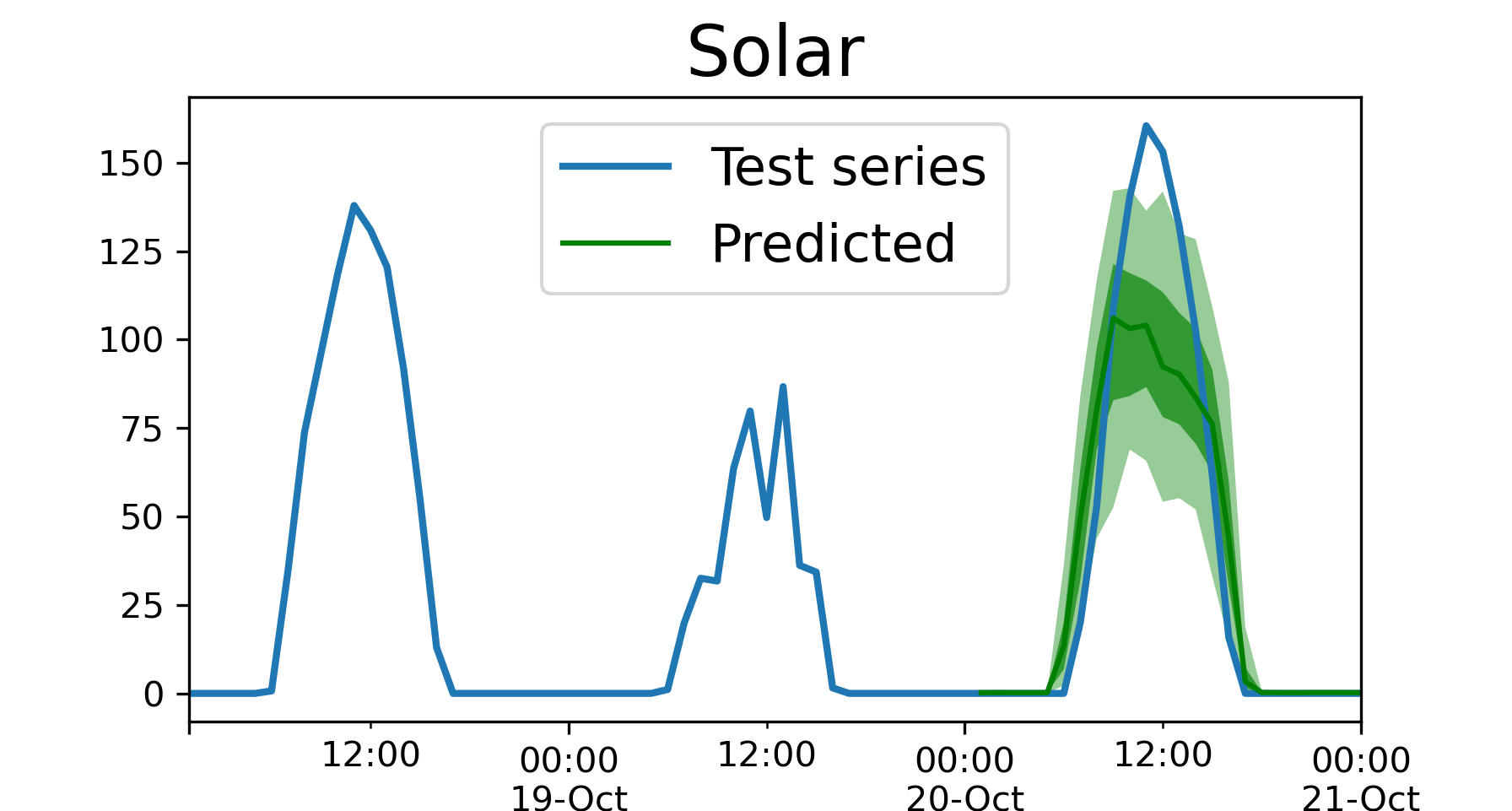}  
       \includegraphics[width=.23\textwidth]{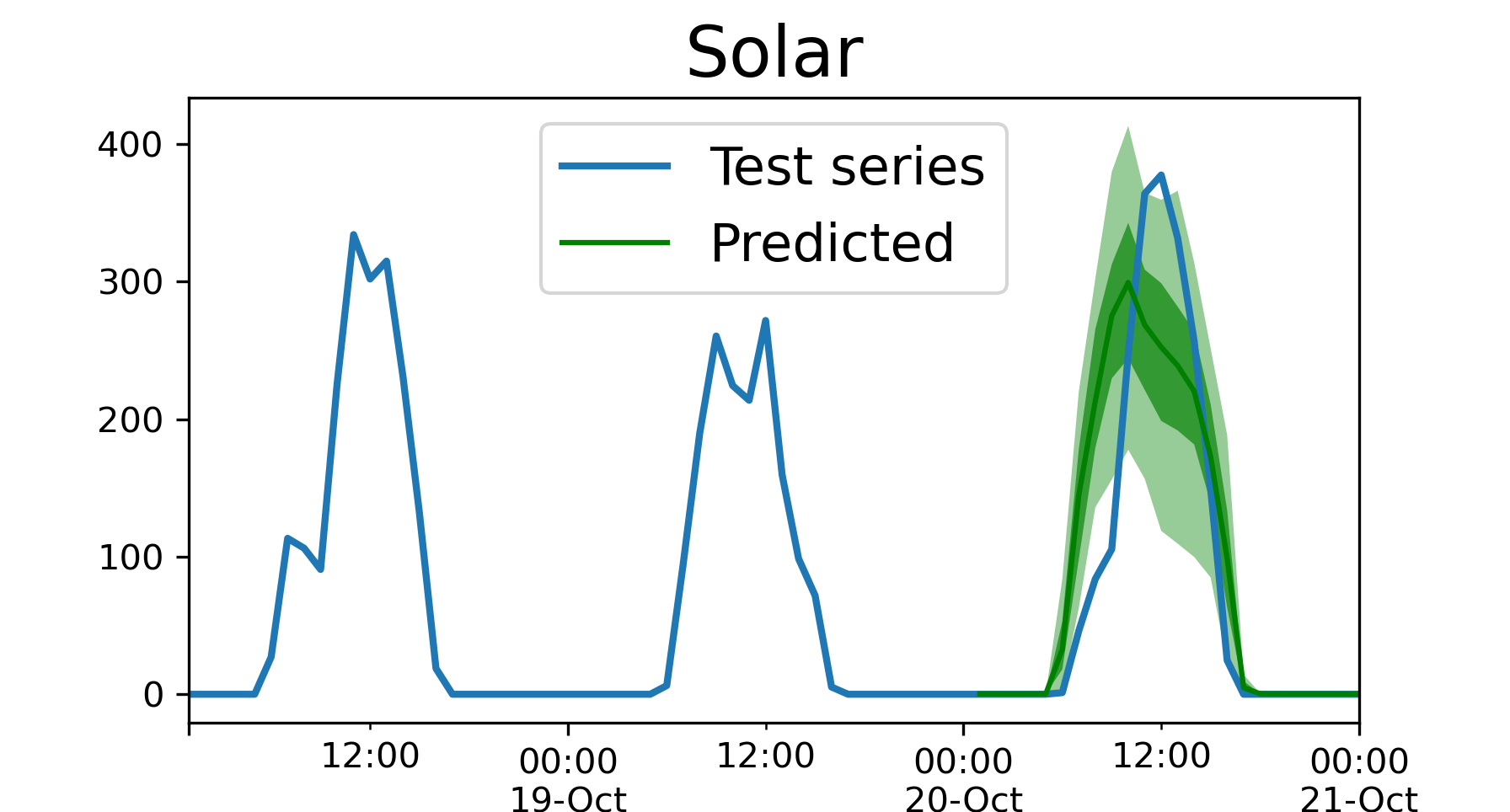}  
      \includegraphics[width=.23\textwidth]{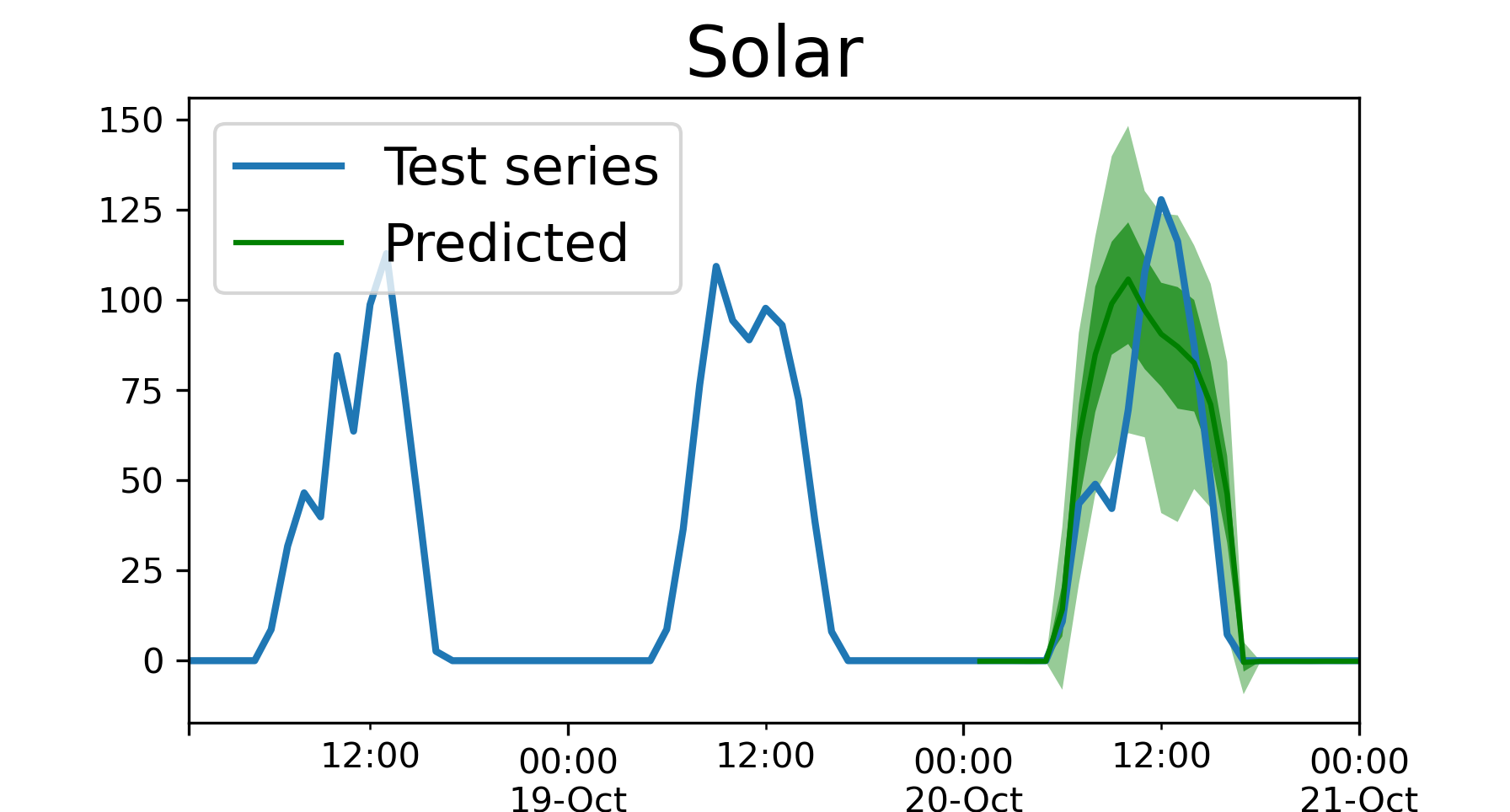}  
            \includegraphics[width=.23\textwidth]{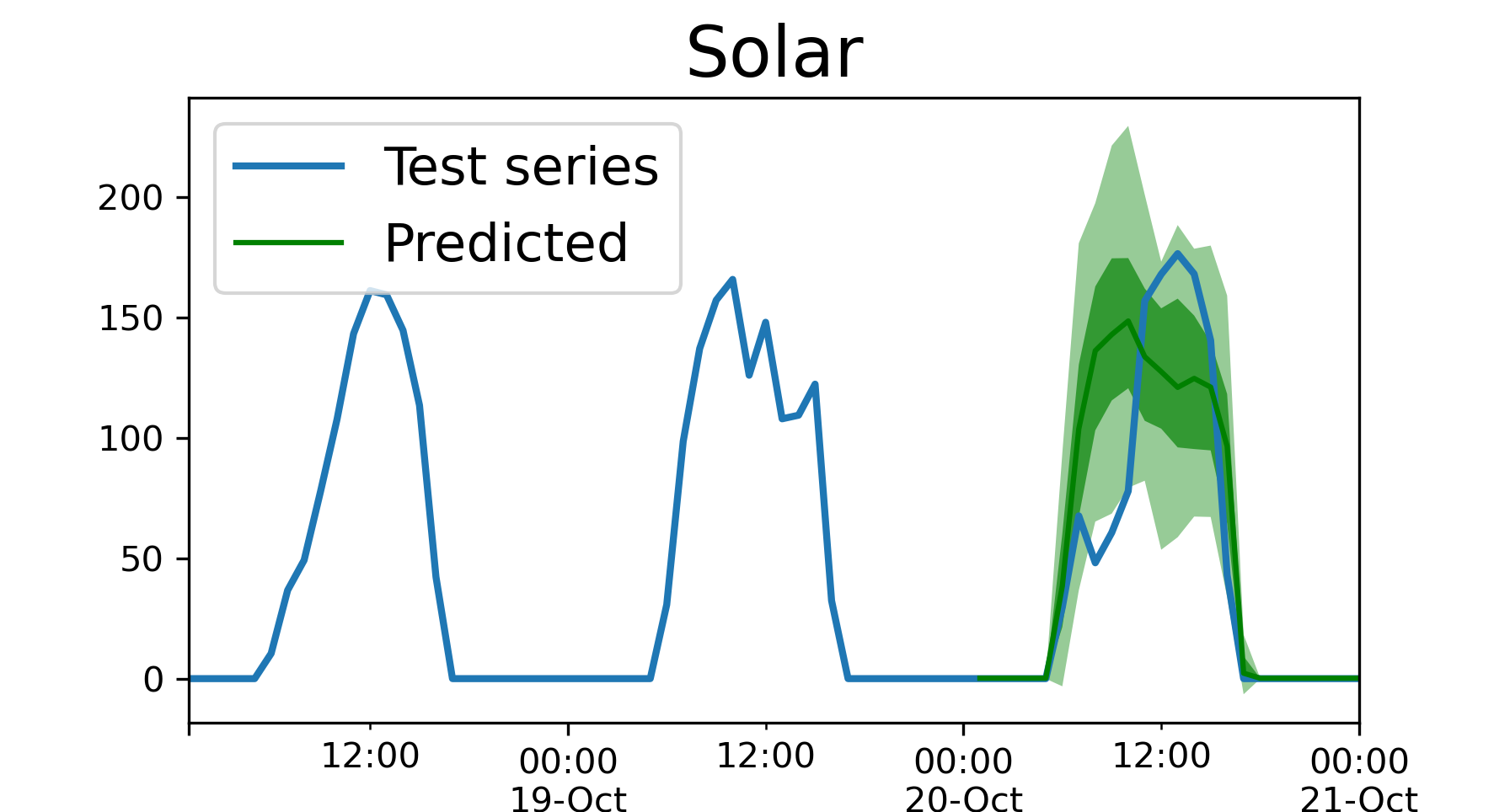}
        \includegraphics[width=.23\textwidth]{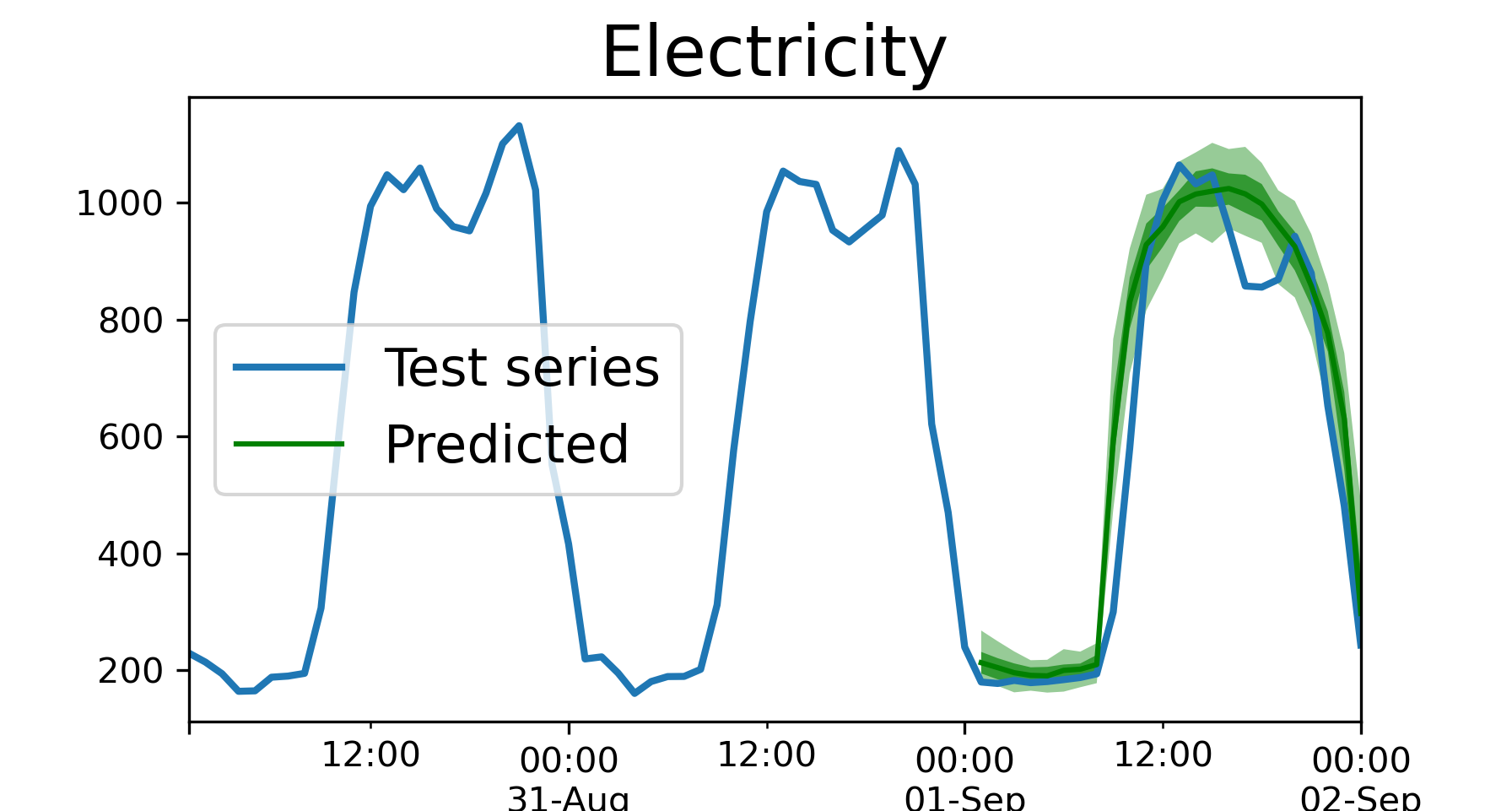}  
       \includegraphics[width=.23\textwidth]{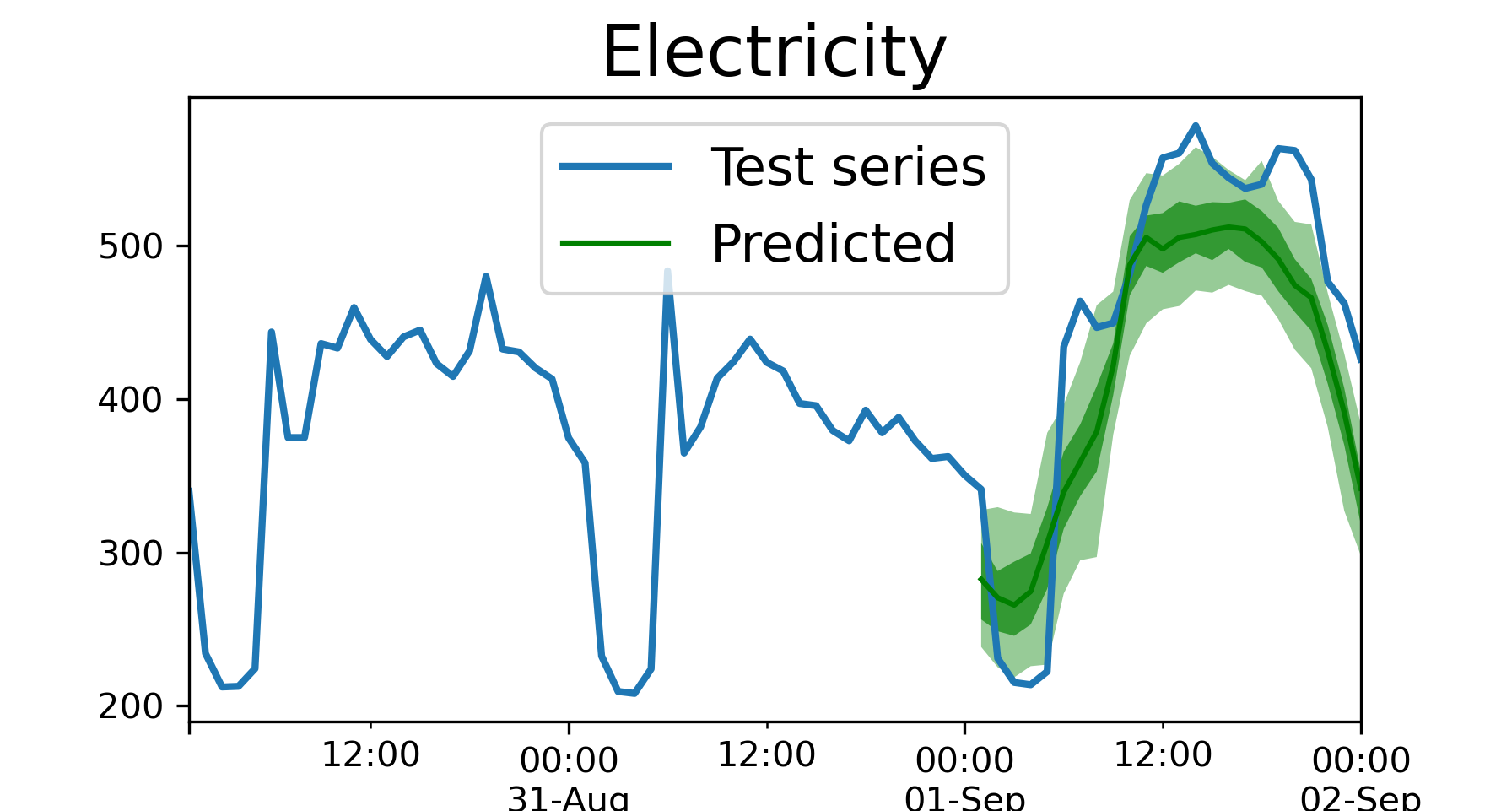}  
      \includegraphics[width=.23\textwidth]{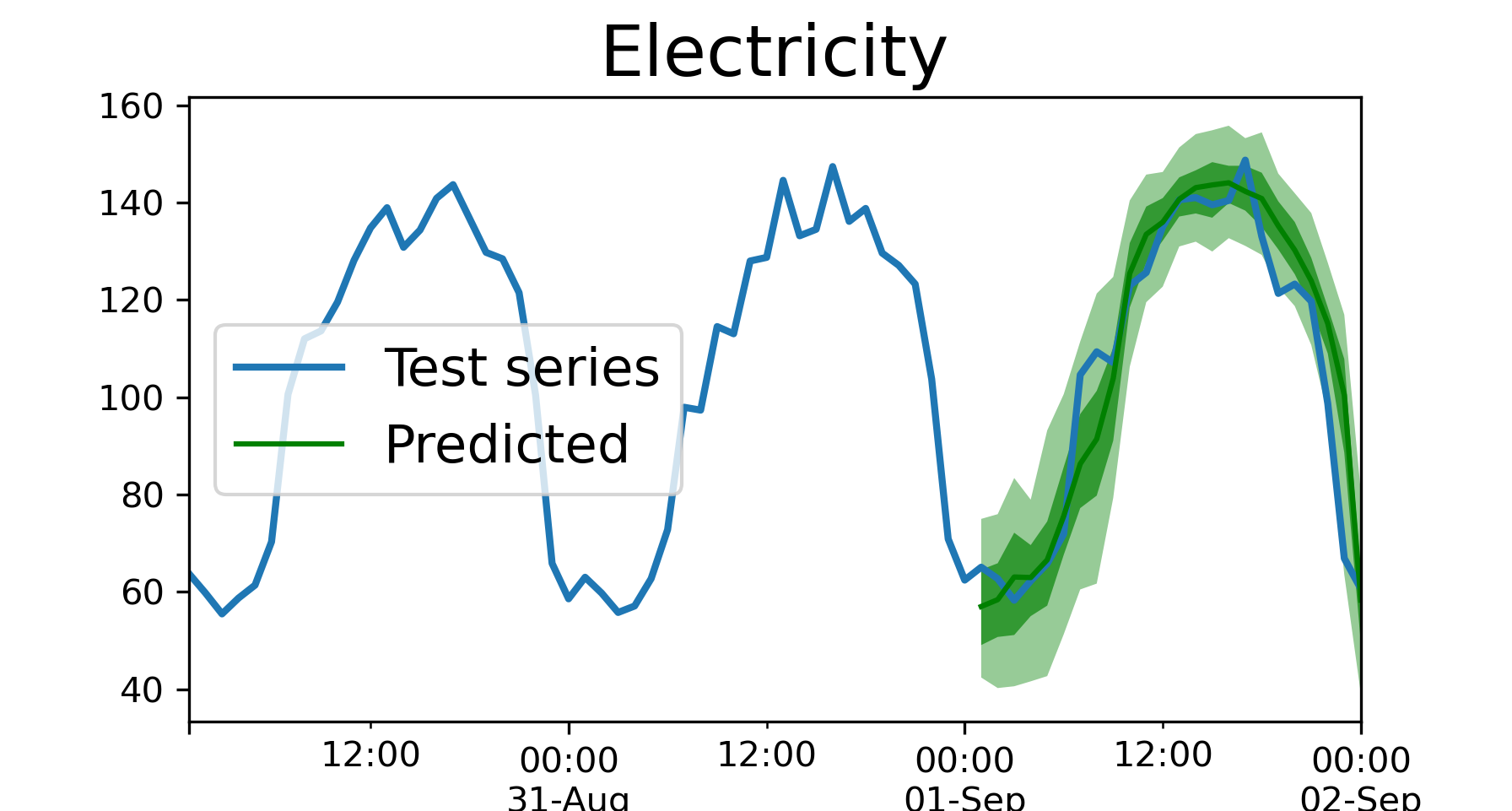}  
            \includegraphics[width=.23\textwidth]{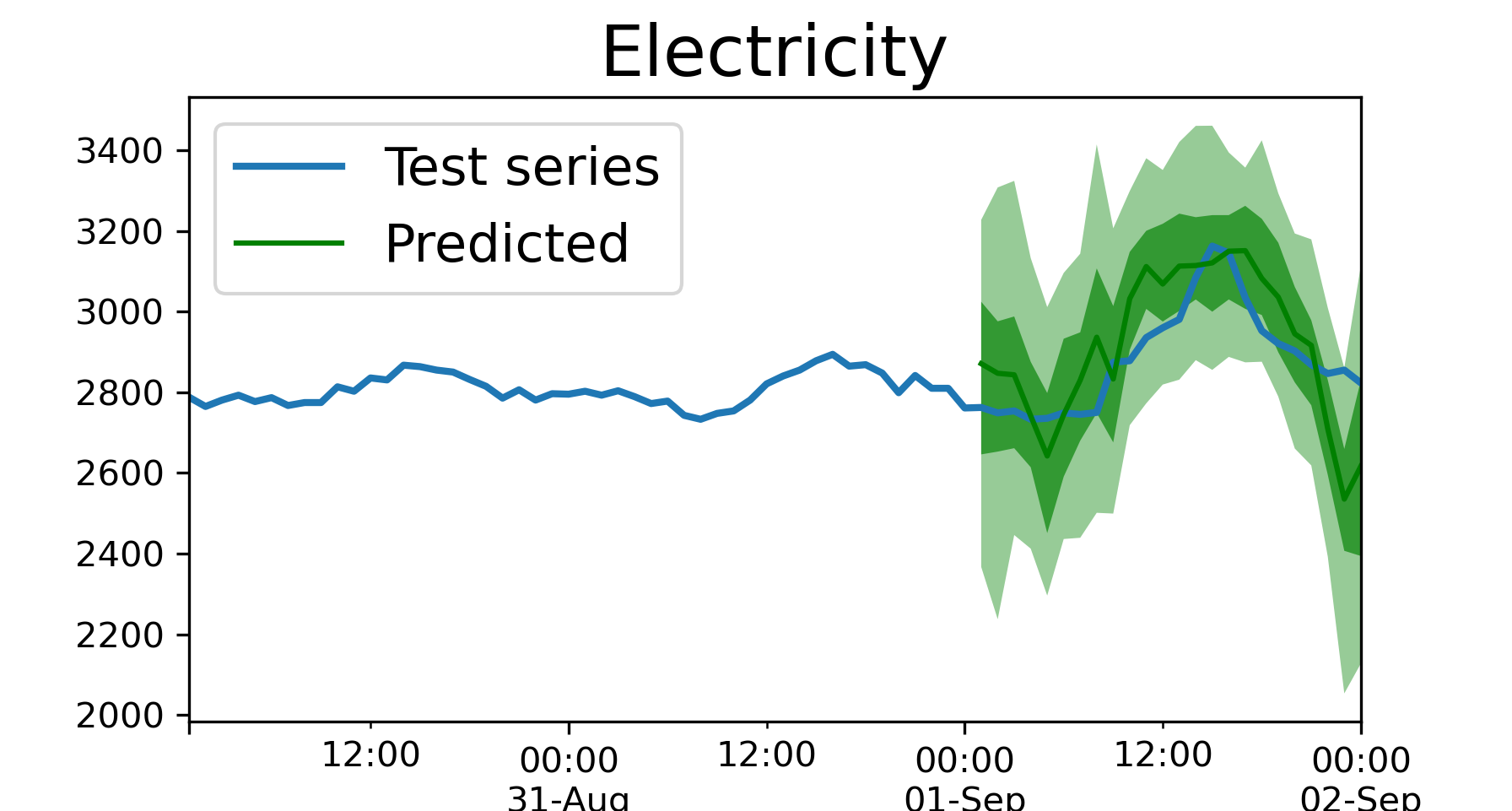}
  \caption{Forecasts generated on Solar (first row) and Electricity (second row) datasets by Temporal Latent Flows. We present predictions the first $4$ dimensions per dataset. Although the datasets are even higher-dimensional, unlike AH\&MS, the signals present strong periodic components.}
   \label{elec}
\end{center}  
\end{figure}

\paragraph{Baselines:} The baselines considered for our experiments include: \textbf{1}: {KVAE}, combines a variational autoencoder with a linear state space model which describes the dynamics. \textbf{2}: {Vec-LSTM-ind-scaling}, models the dynamics via an RNN and outputs the parameters of an independent Gaussian distribution with mean-scaling. \textbf{3}: {Vec-LSTM-lowrank-Copula},  instead parametrizes a low-rank plus diagonal covariance via Copula process. \textbf{4}: {GP-scaling}, unrolls an LSTM with scaling  on each individual time-series before reconstructing the joint distribution via a low-rank Gaussian. \textbf{5}: {GP-Copula}, unrolls an LSTM on each individual time-series and then the joint emission distribution is given by  a low-rank plus diagonal covariance Gaussian copula. \textbf{6}: {TCNF}, uses LSTM to model the temporal conditioning and Masked Autoregressive Flow \citep{papamakarios2017masked} or Real-NVP for the distribution emission model. We also consider \textbf{7}: DeepVAR estimator, which is a multivariate variant of DeepAR \citep{salinas2019deepar}.

\begin{table*}[]
\begin{center}
\scalebox{0.7}{
\begin{tabular}{lccllllll}
\toprule
 & \makecell{ {Solar}} & \makecell{{Electricity}} & \makecell{{Traffic}}  & \makecell{{Taxi}} & \makecell{{Wikipedia}} \\
\midrule
\makecell{{VRNN} \citep{chung2015recurrent}}   & 
$\bm{0.133 \scriptstyle{\pm 0.009}}$ & 
$0.051 \scriptstyle{\pm 0.001}$ & 
${0.181 \scriptstyle{\pm 0.002}}$ & 
$0.139 \scriptstyle{ \pm 0.015}$ &  

$3.400\scriptstyle{\pm 0.322}$ \\ 

\makecell{{KVAE} \citep{fraccaro2017disentangled}}  & 
$0.344\scriptstyle{\pm0.002}$
& $0.051 \scriptstyle{\pm 0.025}$
& $-$
& \bm{$-$}
& $0.099 \scriptstyle{\pm 0.012}$   \\

\makecell{{TCNF LSTM - RealNVP} \citep{rasul2020multi}}  & 
$0.365 \scriptstyle{\pm 0.002}$   &
$0.027 \scriptstyle{\pm 0.001}$ & 
$\bm{0.026 \scriptstyle{\pm 0.001}}$ & 
${0.182 \scriptstyle{\pm 0.013}}$ & 

${0.097 \scriptstyle{\pm 0.024}}$ \\ 

\makecell{{TCNF LSTM - MAF} \citep{rasul2020multi}}
& $0.378 \scriptstyle{\pm 0.032}$ 
& $0.025 \scriptstyle{\pm 0.002}$      
& $\bm{0.028 \scriptstyle{\pm 0.002}}$ 
&  $0.172\scriptstyle{\pm 0.001}$  

& $0.097 \scriptstyle{\pm 0.033}$\\

\makecell{{Vec-LSTM} - {ind-scaling} \citep{salinas2019high}} 

&  $0.391 \scriptstyle{\pm 0.017}$ 
& $0.025 \scriptstyle{\pm 0.001}$
& $0.087 \scriptstyle{\pm 0.041}$ 
& $0.506 \scriptstyle{\pm 0.005}$ 
&  $0.133 \scriptstyle{\pm 0.002}$      \\

\makecell{{Vec-LSTM} -{lowrank-Copula} \citep{salinas2019high}}  
&  $0.319 \scriptstyle{\pm 0.011}$ 
& $0.064 \scriptstyle{\pm 0.008}$ 
& ${0.103 \scriptstyle{\pm 0.006}}$ 
&  $0.326 \scriptstyle{\pm 0.007}$ 

& $0.241 \scriptstyle{\pm 0.033}$    \\

\makecell{ {GP} - {Scaling} \citep{salinas2019high}} 
& $0.368 \scriptstyle{\pm 0.012}$
& \bm{$0.022 \scriptstyle{\pm 0.000}$}
& $0.079 \scriptstyle{\pm 0.000}$

& $0.183 \scriptstyle{\pm 0.395}$
& $1.483 \scriptstyle{\pm 1.034}$  \\

\makecell{{GP} - {Copula} \citep{salinas2019high}}
& $0.337 \scriptstyle{\pm 0.024}$
& $0.024 \scriptstyle{\pm 0.002}$
& $0.078 \scriptstyle{\pm 0.002}$

& $0.208 \scriptstyle{\pm 0.183}$
& \bm{$0.086 \scriptstyle{\pm 0.004}$}  \\

\makecell{{Latent Temporal Flows - Real NVP}}
& \bm{$0.122\scriptstyle{\pm 0.00}$}
&  $\bm{0.022 \scriptstyle{\pm 0.001}}$  
&$0.045 \scriptstyle{\pm 0.002}$    
&  $\bm{0.128 \scriptstyle{\pm 0.015}}$

&  $0.106 \scriptstyle{\pm 0.012}$\\

\makecell{{Latent Temporal Flows - MAF}}  
& $0.217\scriptstyle{\pm 0.001}$  
& \bm{$0.020\scriptstyle{\pm 0.001}$} 
& $0.056 \scriptstyle{\pm 0.002}$  
& $\bm{0.131 \scriptstyle{\pm 0.012}}$  

&  $\bm{0.096 \scriptstyle{\pm 0.017}}$  \\
\bottomrule
\end{tabular}
}

\end{center}
\caption{Test set CRPS comparison (lower is better) of models from V-RNN, GP Copula, TCNF, and our model variants LatTe with Real-NVP, LatTe with-MAF. The mean and standard errors are obtained by re-running each method three times with random initializations. Best performance (top-2) is highlighted in bold. \label{crps2}}
\vspace{-4mm}
\end{table*}

Batch sets are formed by randomly sampling context and adjoining prediction horizon interval sized windows from the complete time-series history of the training data. For the test set, we perform rolling prediction evaluation. Following \citep{salinas2019high}, \textbf{Solar}, \textbf{Electricity}, and \textbf{Traffic}, accuracy is measured on $7$ rolling time windows, for \textbf{Traffic} we use $5$ time windows, and for \textbf{Taxi} $57$ windows are used in order to cover the full test set. We perform rolling prediction evaluation: $24$ time-points per window, last $7$ windows for testing for traffic and electricity, and $14$ per window with last $4$ windows for wiki.

We evaluate forecasting performance starting on equally spaced time points after the last observed training point. Since the marginal continuous ranked probability score (CRPS) cannot assess whether dependencies across time-series are accurately captured, we report the the \emph{Continuous Ranked Probability Score} (CRPS) (CRPS-Sum). This metric is obtained by first summing across the $N$ (ground-truth data and forecasted) time-series, which yields a CDF estimate $\hat F_{\mathrm{sum}}(t)$ for each time point. The results are then averaged over the prediction horizon. We take $100$ samples to estimate the empirical CDF in practice. Table~\ref{crps2} lists the mean CRPS-Sum values averaged over $10$ independent runs with standard deviations and shows that the model sets the new state-of-the-art on most of the benchmark data sets. According to Table~\ref{crps2}, the best performance is achieved by either one of two variants of our proposed model (either Latent Temporal Flows - Real NVP or Latent Temporal Flows - MAF) on Solar, Electricity, and Taxi datasets, and our model’s performance is within top-2 on the Wikipedia dataset.

\section{Discussion}

Personal devices, such as mobile phones and smart watches, include an ever increasing number of sensors that allows for convenient physiological data collection. The abundance of complex wearables data calls for accurate machine learning tools to better forecast and detect changes in the health of individuals. In this paper, we presented a novel method for multivariate forecasting of complex time-series such as vital signs. We have shown that the combination of representation learning, flexible density models, and auto-regressive structures can produce accurate forecasts by modeling dynamics on a low-dimensional subspace. We further analyzed a challenging  practical application of predicting sensor signal trajectories in the AH\&MS dataset.  The proposed approach consistently achieves the best performance over the state-of-the-art methods on publicly available datasets. While here our focus is on time-series forecasting, future work will extend our framework to tasks that require learning over high-dimensional time-series: imputation, interpolation, anomaly detection, and out-of-distribution detection.

\section*{Acknowledgements}

We would like to thank Guillermo Sapiro, Calum MacRae, Rahul Deo, James Kretlow, Shirley Ren, Andrew Miller, and Joseph Futoma for valuable feedback and comments on our manuscript.

\bibliography{citations}

\end{document}